\documentclass[runningheads]{llncs}

\PassOptionsToPackage{table, dvipsnames}{xcolor}
\usepackage{eccv}

\usepackage{eccvabbrv}
\usepackage{cite}

\usepackage{graphicx}
\usepackage{amsmath,amssymb}
\usepackage{cite}
\usepackage{booktabs} 
\usepackage{siunitx}
\usepackage{cuted} 
\usepackage{capt-of}
\usepackage{tabularx}
\usepackage{multirow}
\usepackage{tablefootnote}
\usepackage{threeparttable}

\usepackage{algorithm}
\usepackage{algpseudocode}
\usepackage{etoc}

\usepackage{pifont}

\usepackage{pgfplots}
\pgfplotsset{compat=1.18} 
\usepgfplotslibrary{groupplots}
\usepgfplotslibrary{statistics}
\usepgfplotslibrary{fillbetween}

\definecolor{TUMBlue}{HTML}{0065BD}
\definecolor{TUMDarkBlue}{HTML}{005293}
\definecolor{TUMLightBlue}{HTML}{64A0C8}
\definecolor{TUMOrange}{HTML}{E37222}
\definecolor{TUMGreen}{HTML}{A2AD00}
\definecolor{TUMGray}{HTML}{999999}
\definecolor{TUMLightGray}{HTML}{DAD7CB}

\colorlet{cNuScenes}{TUMOrange}
\colorlet{cAugmented}{TUMDarkBlue}
\colorlet{cCARLA}{TUMBlue}
\colorlet{cCombined}{TUMLightBlue}
\colorlet{HeaderGray}{TUMLightGray!45}
\colorlet{SectionGray}{TUMLightGray!35}

\newcommand{\cmark}{\textcolor{ForestGreen}{\ding{51}}}%
\newcommand{\xmark}{\textcolor{BrickRed}{\ding{55}}}%

\pgfplotscreateplotcyclelist{tumCycle}{%
  {cNuScenes, thick, mark=none},
  {cCARLA,    thick, mark=none},
  {cCombined, thick, mark=none},
  {TUMGreen,  thick, mark=none},
}

\pgfplotsset{
  tumPlot/.style={
    cycle list name=tumCycle,
    width=\linewidth,
    height=0.52\linewidth,
    grid=none,
    major grid style={line width=0.2pt, draw=TUMLightGray},
    minor grid style={line width=0.1pt, draw=TUMLightGray!70},
    tick align=outside,
    tick style={black, line width=0.4pt},
    axis line style={black, line width=0.6pt},
    label style={font=\small},
    tick label style={font=\small},
    legend style={font=\small, draw=none, fill=none, cells={anchor=west}},
    line join=round,
    line cap=round,
  },
}

\usepackage[accsupp]{axessibility}  

\usepackage{hyperref}

\usepackage{orcidlink}

\begin{document}

\title{EgoDyn-Bench: Evaluating Ego-Motion Understanding in Vision-Centric Foundation Models for Autonomous Driving} 

\titlerunning{EgoDyn-Bench}

\author{Finn Rasmus Schäfer\inst{1}\orcidlink{0009-0005-8885-9809} \and 
Yuan Gao\inst{1}\orcidlink{0009-0004-9158-7202} \and
Dingrui Wang \inst{1}\orcidlink{0009-0003-7546-2226} \and 
Thomas Stauner \inst{2}\orcidlink{0000-0003-2669-7195} \and 
Stephan Günnemann\inst{3} \orcidlink{0000-0001-7772-5059} \and
Mattia Piccinini\inst{1}\orcidlink{0000-0003-0457-8777} \and
Sebastian Schmidt\inst{2,3} \orcidlink{0009-0005-4649-1321} \and
Johannes Betz\inst{1}\orcidlink{0000-0001-9197-2849}}

\authorrunning{F.~Schäfer \etal}

\institute{Professorship of Autonomous Vehicle Systems, Technical University of Munich, Munich, Germany \\
\email{finn.schaefer@tum.de}\\
\and
Bayerische Motoren Werke AG, Munich, Germany \\
\and
Data Analytics and Machine Learning Group, Technical University of Munich, Munich, Germany \\
}

\maketitle


\begin{abstract} 
While Vision-Language Models (VLMs) have advanced high-level reasoning in autonomous driving, their ability to ground this reasoning in the underlying physics of ego-motion remains poorly understood. We introduce \textit{EgoDyn-Bench}\footnote{Project page: \href{https://tum-avs.github.io/EgoDyn-Bench-Website/}{TUM-AVS/EgoDyn-Bench-Website}. Code: \href{https://github.com/TUM-AVS/EgoDyn-Bench}{TUM-AVS/EgoDyn-Bench}. Dataset: \href{https://huggingface.co/datasets/fnc1901/EgoDyn-Bench}{fnc1901/EgoDyn-Bench}.}, a diagnostic benchmark for evaluating the semantic ego-motion understanding of vision-centric foundation models. By mapping continuous vehicle kinematics to discrete motion concepts via a deterministic oracle, we decouple a model's internal physical logic from its visual perception. Our large-scale empirical audit spanning 20$+$ models, including closed-source MLLMs, open-source VLMs across multiple scales, and specialized VLAs, identifies a significant \textbf{Perception Bottleneck}: while models exhibit logical physical concepts, they \emph{consistently fail to accurately align them with visual observations}, frequently underperforming classical non-learned geometric baselines. This failure persists across model scales and domain-specific training, indicating a structural deficit in how current architectures couple visual perception with physical reasoning.

We demonstrate that providing explicit trajectory encodings substantially restores physical consistency across all evaluated models, revealing a functional \textbf{disentanglement between vision and language}: ego-motion logic is derived almost exclusively from the language modality, while visual observations contribute negligible temporal signal. This structural finding provides a standardized diagnostic framework and a practical pathway toward physically aligned embodied AI. 

\keywords{Ego-motion \and Physical Reasoning \and Foundation Models} 
\end{abstract}

\section{Introduction}

Classical autonomous driving systems explicitly model ego-motion through estimated physical state variables such as velocity, acceleration, and yaw rate \cite{5940562, Heilmeier02102020, 8917032}. These representations are not auxiliary but fundamental: they ensure that perception, planning, and control remain consistent with the vehicle's underlying dynamics.

Recent advances in vision-centric foundation models, particularly Vision-Language Models (VLMs), propose an alternative paradigm in which high-level reasoning and planning are performed directly from visual observations \cite{10531702, jiang2025surveyvisionlanguageactionmodelsautonomous, sima2025drivelmdrivinggraphvisual, xu2024drivegpt4interpretableendtoendautonomous}. In these approaches, explicit ego-state representations are typically absent, and motion must be inferred implicitly from image sequences.

This shift raises a fundamental question: \emph{Do such models form a physically consistent understanding of ego-motion, or does their visual reasoning remain decoupled from the vehicle's underlying dynamics?}

While current benchmarks primarily assess high-level planning and reasoning tasks~\cite{tian2024drivevlmconvergenceautonomousdriving, qian2024nuscenesqamultimodalvisualquestion, xie2025drivebench}, none verify whether model outputs are physically consistent with the ego-vehicle's own kinematic state, leaving a critical axis of embodied understanding entirely unevaluated.

To address this gap, we introduce \textit{EgoDyn-Bench}, a benchmark for explicitly evaluating ego-motion understanding in vision-centric models. We formulate this as a semantic video question-answering task grounded in physically derived labels, enabling controlled and interpretable assessment of whether model predictions semantically align with the underlying dynamics.
Using this framework, we analyze modern vision-centric models and study the role of explicit dynamic information in their performance. Our results highlight limitations of current alignment based on visual observations and dynamic concepts, and additionally provide a strategy for improving models without retraining.

\noindent Our contributions are as follows:
\begin{itemize}
\item \textbf{Ego-Motion Benchmark:} We introduce \textit{EgoDyn-Bench}, a benchmark that explicitly evaluates ego-motion understanding in vision-centric foundation models, isolating physical consistency from downstream task performance.

\item \textbf{Physically-Grounded Evaluation Framework:} We propose a semantic abstraction and deterministic oracle-based labeling pipeline that maps continuous ego-dynamics to interpretable motion concepts, enabling reproducible and controlled evaluation.

\item \textbf{Large-Scale Empirical Analysis:} Through our audit, we show that current VLMs exhibit a fundamental ``visual grounding deficit'', failing to reliably capture ego-motion from visual input alone despite possessing physically consistent but biased reasoning capabilities. 

\item \textbf{Recovering Grounding via Explicit Dynamics:} We demonstrate that providing textual dynamic state information yields substantial performance gains, providing a pathway to improve physical consistency without the need for expensive retraining.
\end{itemize}

\begin{figure}[t]
  \centering
\includegraphics[width=\textwidth]{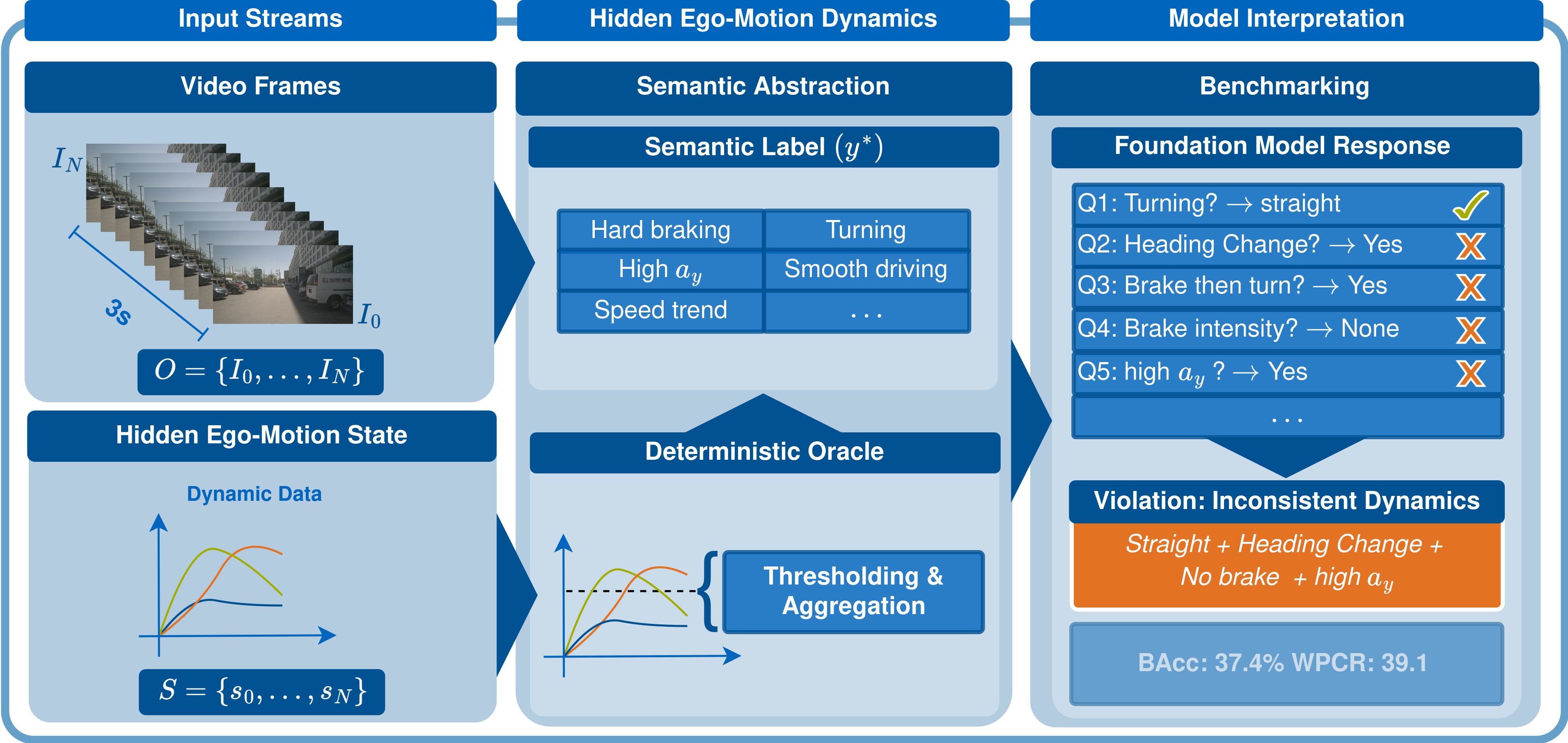}
  \captionof{figure}{\textbf{EgoDyn-Bench Overview.} Continuous kinematic states $S$ are mapped to semantic labels via a deterministic oracle to define a VideoQA task over visual observations $O$. Models are evaluated on their ability to infer motion dynamics through semantic, temporal, and physical consistency (\textit{WPCR}) metrics.}
  \label{fig:teaser}
\end{figure}

\section{Related Work}
Existing evaluation frameworks for vision-centric foundation models in the autonomous driving domain can be grouped into four primary paradigms: (i) logical reasoning and decision interpretability, (ii) spatial intelligence, (iii) numerical trajectory forecasting and closed-loop driving, and (iv) physical reliability audits. Our work targets a missing axis across all four: \emph{whether predicted decisions are consistent with the underlying physical concepts of ego-motion over time}.

\textbf{Standard Autonomous Driving Benchmarks and Metrics}
Classical autonomous driving evaluation spans both dataset-based and simulator-based protocols. Large-scale datasets such as nuScenes~\cite{caesar2020nuscenesmultimodaldatasetautonomous} and Waymo~\cite{waymo} support open-loop evaluation of perception and trajectory prediction, typically using displacement-based metrics (\eg, ADE/FDE). In contrast, closed-loop benchmarks and simulators such as nuPlan~\cite{karnchanachari2024learningbasedplanningthenuplanbenchmark} and CARLA~\cite{dosovitskiy2017carlaopenurbandriving} evaluate planning and control through metrics such as route completion and time-to-collision. While these protocols effectively assess task-level completion, they abstract away the underlying physical reasoning. As a result, models might achieve high performance through spurious visual correlations rather than genuine kinematic understanding, posing significant risks to downstream generalization and safety.

\textbf{Logical Reasoning and Decision Interpretability (LR)}
Frameworks such as DriveLM \cite{sima2025drivelmdrivinggraphvisual} and Reason2Drive \cite{nie2024reason2driveinterpretablechainbasedreasoning} evaluate reasoning through structured or interpretable decision outputs. These approaches represent behavior as discrete linguistic instructions, focusing on semantic plausibility. Because these linguistic instructions are inherently discrete and lack continuous temporal constraints, they cannot guarantee that a sequence of decisions will result in, or be grounded in, a kinematically feasible maneuver. Our setting addresses this by requiring reasoning grounded in temporally continuous motion rather than isolated semantic descriptions.

\textbf{Object-Centric Spatial Intelligence.} 
Recent benchmarks like Ego3D-Bench~\cite{gholami2025spatialreasoningvisionlanguagemodels} and RADAR~\cite{chen2026radarbenchmarkingvisionlanguageactiongeneralization} evaluate spatial relations and volumetric overlap of external objects. In contrast, ego-motion understanding is fundamentally self-referential, requiring the grounding of the agent’s own motion state within the temporal visual stream. \textit{EgoDyn-Bench} addresses this gap by providing a structured diagnostic to evaluate whether a model's high-level semantic interpretation of its own movement is accurately anchored in physical concepts.

\textbf{Trajectory Forecasting and Control.}
Standard benchmarks like Argoverse~\cite{wilson2023argoverse2generationdatasets}, ScenePilot-Bench~\cite{wang2026scenepilotbenchlargescaledatasetbenchmark}, and EgoTraj-Bench~\cite{liu2025egotrajbenchrobusttrajectoryprediction} evaluate motion via displacement-based metrics. However, spatial accuracy does not guarantee kinematic feasibility or compliance with underlying physical concepts. Instead of assessing motion generation, \textit{EgoDyn-Bench} provides an isolated diagnostic of the model’s intrinsic high-level physical understanding, evaluating whether its semantic interpretations are accurately anchored in continuous kinematic constraints.

\textbf{General Physics Audits.} 
Benchmarks like DriveBench~\cite{xie2025drivebench} expose ``text-only resilience,'' where models rely on language priors rather than visual grounding. Physics audits such as QuantiPhy~\cite{puyin2025quantiphyquantitativebenchmarkevaluating} and Morpheus~\cite{zhang2025morpheusbenchmarkingphysicalreasoning} show that models struggle with general conservation laws and external object collisions. In contrast, \textit{EgoDyn-Bench} isolates the understanding of embodied kinematics, testing whether models can infer their own mechanically valid motion states directly from sequential visual streams.

\textbf{Limitations of Existing Approaches}
To summarize, current evaluation paradigms fracture the driving problem along a consistent fault line: classical approaches, including optical flow, visual odometry, and displacement-based trajectory metrics, offer rigorous geometric tracking but no semantic understanding, while foundation model benchmarks evaluate high-level reasoning from visual snapshots without any grounding in the vehicle's own kinematic state. \textit{EgoDyn-Bench} bridges this division by formally testing whether vision-centric semantic predictions satisfy the kinematic concepts related to continuous ego-motion. Our analysis reveals that model failures stem from misalignment between visual observations and physical motion concepts, not from an absence of physical 
reasoning capacity.

\begin{table}[ht]
\centering
\caption{Comparison of VLM evaluation benchmarks in autonomous driving. Existing benchmarks evaluate external environments or abstract actions, but fail to assess the agent's internal kinematics. \textit{EgoDyn-Bench} isolates this missing self-referential axis. \textbf{OCS}: Object-Centric Spatial; \textbf{CLT}: Closed-Loop \& Trajectory; \textbf{LR}: Logical Reasoning; \textbf{GPA}: General Physics Audits; \textbf{KEM}: Kinematic Ego-Motion.}
\label{tab:related_work}
\scriptsize
\renewcommand{\arraystretch}{1.05}
\setlength{\tabcolsep}{6pt} 

\begin{tabular}{@{} l c c c c c @{}} 
\toprule
\textbf{Benchmark} & \textbf{OCS} & \textbf{CLT} & \textbf{LR} & \textbf{GPA} & \textbf{KEM} \\
\midrule

\rowcolor{SectionGray}
\multicolumn{6}{c}{\textbf{Classical \& Trajectory Benchmarks}} \\
\midrule
\textit{nuScenes-QA}~\cite{qian2024nuscenesqamultimodalvisualquestion} & \cmark & \xmark & \xmark & \xmark & \xmark \\
\textit{nuPlan}~\cite{karnchanachari2024learningbasedplanningthenuplanbenchmark} & \cmark & \cmark & \xmark & \xmark & \xmark \\
\textit{ScenePilot / EgoTraj}~\cite{wang2026scenepilotbenchlargescaledatasetbenchmark} & \xmark & \cmark & \xmark & \xmark & \xmark \\

\midrule
\rowcolor{SectionGray}
\multicolumn{6}{c}{\textbf{Vision-Language \& Reasoning Benchmarks}} \\
\midrule
\textit{DriveLM / Reason2Drive}~\cite{sima2025drivelmdrivinggraphvisual} & \xmark & \xmark & \cmark & \xmark & \xmark \\
\textit{Ego3D / RADAR}~\cite{gholami2025spatialreasoningvisionlanguagemodels} & \cmark & \xmark & \cmark & \xmark & \xmark \\
\textit{DriveBench / QuantiPhy}~\cite{xie2025drivebench} & \xmark & \xmark & \xmark & \cmark & \xmark \\

\midrule
\rowcolor{SectionGray}
\multicolumn{6}{c}{\textbf{Proposed Benchmark}} \\
\midrule
\textbf{\textit{EgoDyn-Bench} (Ours)} & \xmark & \xmark & \xmark & \xmark & \cmark \\
\bottomrule
\end{tabular}
\end{table}


\section{EgoDyn-Bench}
To examine the kinematic motion understanding of foundation models, we introduce \textit{EgoDyn-Bench}. Our benchmark is the first to evaluate whether vision-centric models can infer physically consistent ego-motion concepts from visual observations. A comparison to existing frameworks is shown in \Cref{tab:related_work}.
Unlike prior object-centric or regression-based benchmarks, we isolate \textit{self-referential motion understanding} as a semantic reasoning problem grounded in vehicle kinematics. Our benchmark consists of: (i) a task formulation mapping visual inputs to semantic motion concepts, (ii) a dataset of real-world and augmented driving sequences, and (iii) a reproducible labeling pipeline deriving ground-truth from physical signals. \Cref{fig:teaser} provides a conceptual overview.

\subsection{Problem Formulation}\label{subsec:problem-formulation}

\textbf{Goal.} We formulate visual ego-motion understanding as a semantic question-answer task: given a \textit{visual observation sequence} and a \textit{natural language query} regarding the vehicle's movement, a model must produce a \textit{semantic response} grounded in the underlying motion. By shifting from raw state regression to a linguistic interface, we evaluate a model's ability to extract functional physical concepts from observations rather than simply regressing numerical values.  

\noindent\textbf{Input.} Formally, the model receives a sequence of visual observations 
\[
    O = \{I_0, \dots, I_N\},
\] 
sampled at frequency $f_{cam}$ over a temporal window of duration $\tau$. This is paired with a natural language query $P$ that specifies a motion-related property (\eg, turning direction or braking behavior). Following established protocols for trajectory forecasting and planning~\cite{karnchanachari2024learningbasedplanningthenuplanbenchmark, wilson2023argoverse2generationdatasets}, we utilize fixed-length segments where $\tau = \qty{3}{\second}$. This duration provides sufficient context for characteristic maneuvers to unfold as observable spatio-temporal changes while remaining brief enough to isolate distinct semantic behaviors and avoid confounding scene transitions. While our experiments utilize 3-second clips, the formulation remains flexible across varying temporal horizons.

\noindent\textbf{Kinematics.} The ego-vehicle's motion is characterized by a physical state sequence $S = \{s_0, \dots, s_N\}$, recorded from onboard sensors or simulated ground-truth during data collection and strictly withheld from the model during inference.
\[
s_t = \left[v_t, a_t, j_t,  \omega_t, \theta_t\right]^\top
\]
captures the vehicle's speed $v_t$, longitudinal acceleration $a_t$, jerk $j_t$, yaw rate $\omega_t$, and heading $\theta_t$. Rather than regressing these exact numerical states, our formulation probes whether models grasp the functional physics of ego-motion, evaluating the semantic implications of these kinematics rather than precise estimation.

\noindent\textbf{Evaluation.} We define a vision-centric foundation model $\mathcal{F}_\theta$ that maps the visual and textual inputs to a predicted semantic response:
\[
\mathcal{F}_\theta(O, P) \rightarrow \hat{R},
\]
where $\hat{R}$ is an answer selected from a predefined semantic space (binary or multiple-choice options). To enable objective evaluation, we define a deterministic \textbf{oracle} $\mathcal{G}$:
\[
\mathcal{G}(S, P) \rightarrow R^*,
\]
which maps the physical state sequence $S$ and query $P$ to a ground-truth semantic answer $R^*$. This ensures that labels are derived directly from measurable kinematics rather than subjective human annotation. Given the short horizon $\tau$, sensor drift is negligible, and our sensitivity analysis confirms that model rankings remain robust to variations in the oracle's thresholds (see Supplementary).

\subsection{Dataset Construction}
To enable controlled evaluation of ego-motion understanding, our data generation pipeline must satisfy three requirements: broad coverage of ego-dynamics, access to withheld physically grounded motion signals for oracle annotation, and a controllable distribution of motion regimes. To achieve this, \textit{EgoDyn-Bench} employs a hybrid approach: we utilize \textit{nuScenes}~\cite{caesar2020nuscenesmultimodaldatasetautonomous} for real-world driving sequences and augment underrepresented motion regimes using targeted simulations in CARLA~\cite{dosovitskiy2017carlaopenurbandriving}, driven by CommonRoad scenarios~\cite{Klischat2019b} and an adaptable motion planner~\cite{frenetix} according to~\cite{gao2026stylevladrivingstyleawarevision}.

\noindent\textbf{Distribution Balancing \& Data Curation.}
Real-world datasets, like \eg, \textit{nuScenes}, provide authentic visual and kinematic synchronization but are inherently biased toward low-dynamic, routine driving. To ensure a physically comprehensive distribution across longitudinal and lateral profiles, we explicitly control the benchmark's statistics through a structured four-stage curation pipeline. First, \textit{dynamic characterization} identifies the natural low-dynamic bias of the real-world logs. Next, \textit{dynamic mining} extracts rare but informative maneuvers directly from \textit{nuScenes}. To balance the remaining underrepresented regions, such as emergency braking or high lateral acceleration, we employ \textit{targeted augmentation}, injecting dynamically diverse trajectories simulated in CARLA. Finally, all sequences undergo human \textit{validation} to verify the extracted motion signals and labeling rules. Detailed curation thresholds are provided in the Supplementary Material. \Cref{fig:dataset_augmentation_combined} illustrates how this mixture effectively corrects spatial and dynamic biases.

\noindent\textbf{Domain Alignment.} 
To mitigate the visual domain gap between simulated and real-world sequences, we apply a photometric style transfer model (NVIDIA Cosmos Transfer~2~\cite{nvidia2025cosmostransfer1conditionalworldgeneration}) to the CARLA scenarios. Because our benchmark evaluates motion understanding rather than photometric fidelity, this alignment strictly prioritizes the preservation of geometric and kinematic cues over exact visual realism. We verify the recoverability of these essential motion signals across domains via geometric baselines, detailed in \Cref{subsec:domain_gap}.

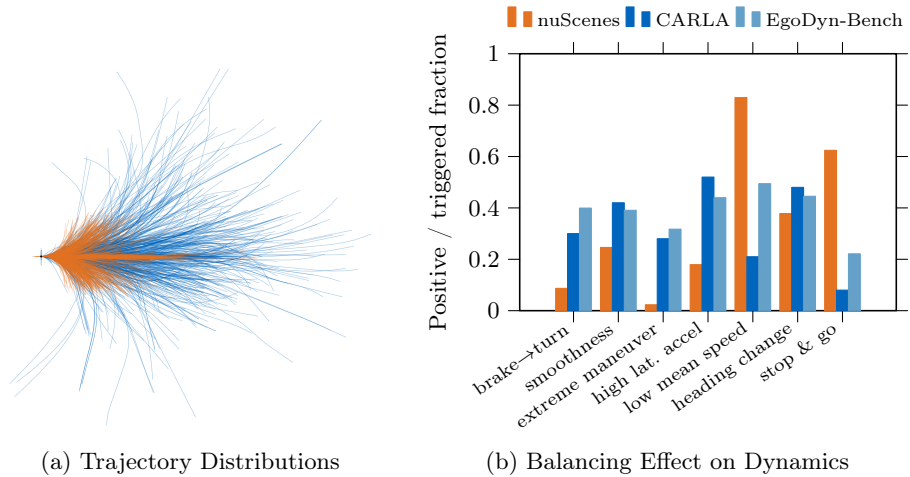
\begin{figure}[t]
    \centering
    
    \begin{minipage}[b]{0.42\linewidth}
        \centering
        \includegraphics[width=\linewidth]{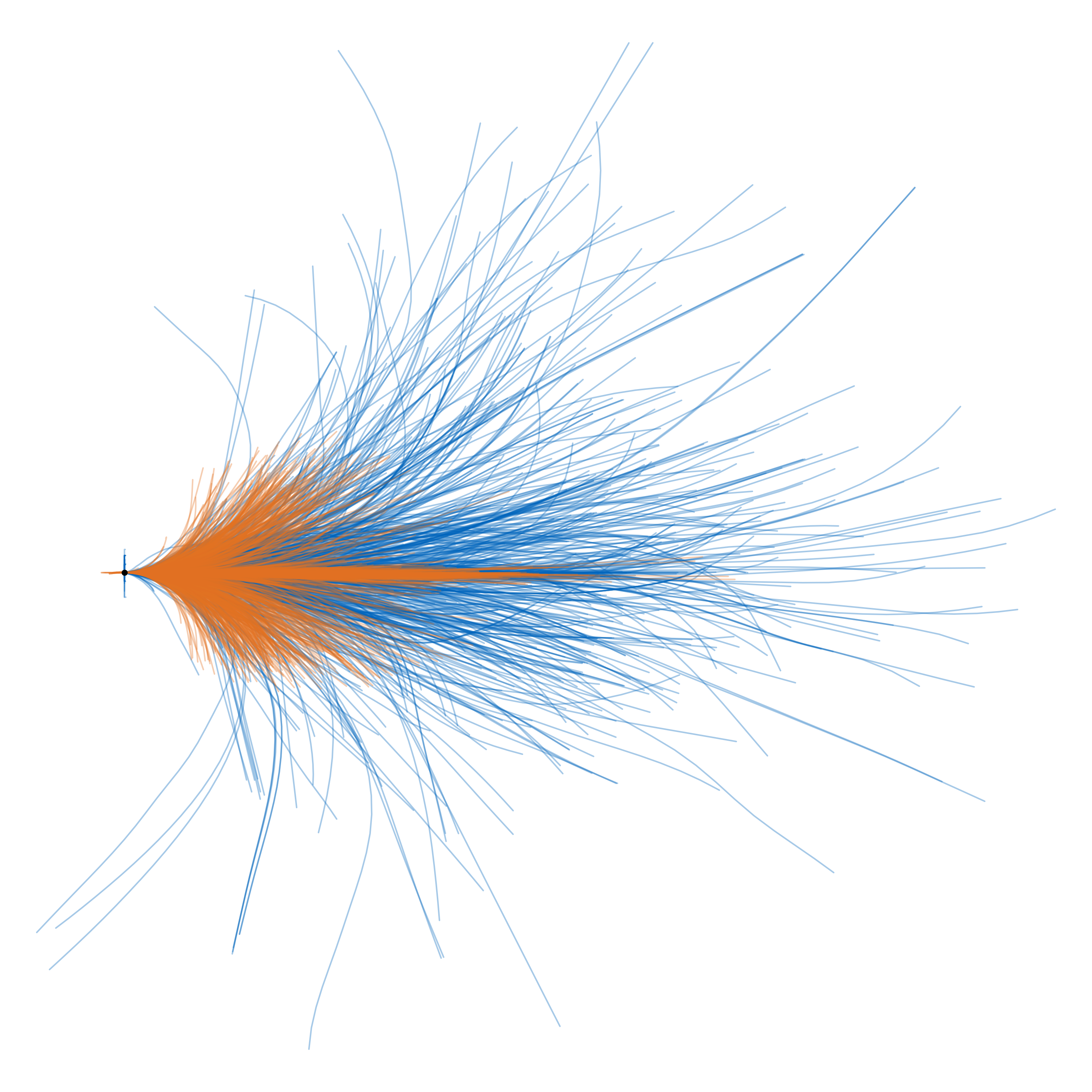}
        \centerline{\small (a) Trajectory Distributions}
    \end{minipage}
    \hfill
    \begin{minipage}[b]{0.54\linewidth}
        \centering
        \begin{tikzpicture}
        \begin{axis}[
            tumPlot,
            bar width=4.5pt, 
            ybar=0pt,        
            width=\linewidth,
            height=5.0cm,
            ymin=0, ymax=1,
            ylabel={Positive / triggered fraction},
            symbolic x coords={
                brake\_then\_turn,
                driving\_smoothness,
                extreme\_maneuver,
                high\_lateral\_accel,
                mean\_speed\_low,
                significant\_heading\_change,
                stop\_and\_go
            },
            xtick=data,
            xticklabels={
                brake$\rightarrow$turn,
                smoothness,
                extreme maneuver,
                high lat.\ accel,
                low mean speed,
                heading change,
                stop \& go
            },
            x tick label style={
                rotate=35,
                anchor=east,
                align=right,
                font=\scriptsize 
            },
            enlarge x limits=0.2, 
            legend style={at={(0.5,1.05)}, anchor=south, legend columns=3, font=\scriptsize},
        ]
        
        \addplot+[draw=cNuScenes, fill=cNuScenes] coordinates {
            (brake\_then\_turn,0.0869)
            (driving\_smoothness,0.2461)
            (extreme\_maneuver,0.0230)
            (high\_lateral\_accel,0.1791)
            (mean\_speed\_low,0.8293)
            (significant\_heading\_change,0.3780)
            (stop\_and\_go,0.6241)
        };
        
        \addplot+[draw=cCARLA, fill=cCARLA] coordinates {
            (brake\_then\_turn,0.3000)
            (driving\_smoothness,0.4200)
            (extreme\_maneuver,0.2800)
            (high\_lateral\_accel,0.5200)
            (mean\_speed\_low,0.2100)
            (significant\_heading\_change,0.4800)
            (stop\_and\_go,0.0800)
        };
        
        \addplot+[draw=cCombined, fill=cCombined] coordinates {
            (brake\_then\_turn,0.3993)
            (driving\_smoothness,0.3907)
            (extreme\_maneuver,0.3173)
            (high\_lateral\_accel,0.4397)
            (mean\_speed\_low,0.4943)
            (significant\_heading\_change,0.4453)
            (stop\_and\_go,0.2213)
        };
        
        \legend{nuScenes, CARLA, EgoDyn-Bench}
        \end{axis}
        \end{tikzpicture}
        \centerline{\small (b) Balancing Effect on Dynamics}
    \end{minipage}
    
    \caption{\textbf{Effect of Dataset Augmentation.} \textbf{(a)} Spatial coverage of \textit{nuScenes} (\textcolor{TUMOrange}{orange}) vs. \textit{CARLA-derived} scenarios (\textcolor{TUMBlue}{blue}). CARLA expands the state-space to include complex maneuvers required for robust benchmarking. \textbf{(b)} Positive label fractions for representative questions. \textit{EgoDyn-Bench} corrects the low-dynamic bias of nuScenes by injecting dynamically augmented synthetic sequences.}
    \label{fig:dataset_augmentation_combined}
\end{figure}

\subsection{Semantic Abstraction and Label Generation}\label{subsec:semantic-abstraction}

We formalize discrete driving maneuvers by applying a deterministic thresholding scheme to the continuous state $S$. These thresholds are calibrated on the dataset distribution and cross-verified against standard automotive kinematics literature~\cite{8933492, 6083078, 6856461} to ensure physical plausibility. Thresholds are used exclusively by the oracle $\mathcal{G}$ to derive ground-truth labels and are never disclosed to evaluated models, which must infer semantic motion concepts from visual observations alone. A full sensitivity analysis under uniform threshold perturbation by a factor $\alpha \in [0.5, 1.5]$, measured via Kendall's $\tau$~\cite{10.1093/biomet/30.1-2.81}, confirms stable model rankings ($\tau > 0.9$) across all perturbation levels. This design ensures robustness and generalization. The semantic definitions can be adapted to new domains or specific research requirements by adjusting the threshold parameters. 

\noindent\textbf{Semantic Categories.} We evaluate 14 distinct question categories within a unified prompt template, spanning two complementary reasoning dimensions: (i) \emph{direct dynamics}, which probe instantaneous or aggregated motion properties such as speed regime, braking intensity, lateral acceleration, and driving smoothness; and (ii) temporal comparative, which require the model to reason about the ordering or co-occurrence of events across the clip.\footnote{Full question prompts, answer options, and labeling rules with calibrated thresholds are provided in the supplementary material.}

\subsection{Evaluation Metrics}\label{subsec:metrics}

We evaluate ego-motion understanding by treating the model’s natural language responses as semantic reasoning over discrete motion concepts. Let $y_i \in \mathcal{Y}$ denote the oracle label for a query $P_i$ on clip $i$, and $\hat{y}_i$ the predicted label obtained by mapping the model response $\hat{R}_i$ to the label space $\mathcal{Y}$ via a deterministic parser. Our metrics are designed to distinguish between a model's ability to extract information from pixels (correctness) and its ability to maintain logically sound internal physical reasoning (consistency). 

\noindent\textbf{Semantic Correctness.} 
While \textit{EgoDyn-Bench} is balanced across question categories to avoid dataset-driven bias, models often exploit internal linguistic priors rather than grounding responses in visual content~\cite{goyal2017makingvvqamatter}. To ensure performance reflects genuine physical reasoning, we utilize \textbf{Balanced Accuracy} and \textbf{Macro-F1}. Balanced Accuracy (BAcc) 
is defined as the mean of class-wise recalls:
\begin{equation}
\mathrm{Bal.\ Acc.} = \frac{1}{|\mathcal{Y}|} \sum_{c \in \mathcal{Y}} \frac{1}{N_c} \sum_{i=1}^{N} 1[\hat{y}_i = c \land y_i = c],
\end{equation}
where $N_c$ is the number of ground-truth instances for class $c$. For temporal queries that compare event ordering, we report \textbf{Temporal Accuracy}, the fraction of correctly predicted temporal orderings. 

\noindent\textbf{Weighted Physics Consistency Rate (WPCR).} Beyond correctness, we introduce WPCR to diagnose the internal coherence of model predictions. We define a set of Boolean physical constraints $\mathcal{R}=\{r_m\}$, as reported in \Cref{tab:WPCR}. Importantly, WPCR is not a measure of accuracy against the ground truth, but a diagnostic of internal physical coherence, assessing whether a model's collective answers for a single clip ($\tau=\qty{3}{\second}$) satisfy the physics of motion.

\begin{table}[ht]
\centering
\caption{\textbf{WPCR Kinematic Constraints.} Boolean implication rules ($A \Rightarrow B$) used to compute WPCR, each evaluated on a single clip.}
\label{tab:WPCR}
\scriptsize
\renewcommand{\arraystretch}{1.05}
\setlength{\tabcolsep}{2pt}

\begin{tabular}{@{} c p{0.78\linewidth} @{}}
\toprule
\textbf{Rule} & \textbf{Boolean Implication ($A \Rightarrow B$)} \\
\midrule

\rowcolor{SectionGray}
\multicolumn{2}{c}{\textbf{Heading / Lateral Dynamics (Hard)}} \\
\midrule
$R_1$ & Heading Change = Yes $\Rightarrow$ Turn Direction $\neq$ Straight \\
$R_2$ & High Lateral Acceleration = Yes $\Rightarrow$ Turn Direction $\neq$ Straight \\
$R_3$ & Turn Direction = Straight $\Rightarrow$ No Significant Heading Change \\
$R_4$ & Turn Direction = Straight $\Rightarrow$ No High Lateral Acceleration \\

\midrule

\rowcolor{SectionGray}
\multicolumn{2}{c}{\textbf{Speed Regime / Mean Speed (Hard)}} \\
\midrule
$R_5$ & Speed Regime = Highway $\Rightarrow$ Mean Speed is Not Low \\
$R_6$ & Speed Regime = Stopped $\Rightarrow$ Mean Speed is Low \\
$R_7$ & Speed Regime = Stopped $\Rightarrow$ Speed Trend $\neq$ Accelerating \\

\midrule

\rowcolor{SectionGray}
\multicolumn{2}{c}{\textbf{Brake-then-Turn Compound (Hard)}} \\
\midrule
$R_8$ & Brake-then-Turn = Yes $\Rightarrow$ Braking Intensity $\neq$ None \\
$R_9$ & Brake-then-Turn = Yes $\Rightarrow$ Turn Direction $\neq$ Straight \\

\midrule

\rowcolor{SectionGray}
\multicolumn{2}{c}{\textbf{Stop-and-Go (Hard)}} \\
\midrule
$R_{10}$ & Stop-and-Go = Yes $\Rightarrow$ Speed Regime $\neq$ Stopped \\

\bottomrule
\end{tabular}
\end{table}

To prevent models from achieving high consistency scores by simply avoiding committed predictions, we weight each clip's consistency contribution by the fraction of rules it triggers. Let $\mathcal{C}$ denote the set of evaluated 
clips, $T_c$ the number of applicable rules and $V_c$ the number of violations for clip $c$:
\begin{equation}
\mathrm{WPCR} = \frac{1}{|\mathcal{C}|} \sum_{c \in \mathcal{C}} \left( 1[V_c = 0 \land T_c > 0] \cdot \frac{T_c}{|\mathcal{R}|} \right).
\end{equation}

Hard Boolean implications are a deliberate design choice: ego-motion concepts are physically discrete and mutually exclusive, leaving no meaningful notion 
of partial correctness. A soft metric would mask systematic reasoning failures behind gradual penalty curves, obscuring the architectural deficits this 
benchmark is designed to expose.

We additionally report \textbf{Physics Coverage (PCov)} as 
$\frac{1}{|\mathcal{C}|} \sum_{c \in \mathcal{C}} \frac{T_c}{|\mathcal{R}|}$, representing the mean fraction of physical constraints triggered per clip. A low PCov indicates that few rules are triggered per clip, which would make a high WPCR score uninformative. High PCov confirms that the consistency rules are actively exercised across the benchmark.

\section{Experiments}

We evaluate the ability of vision-centric foundation models to infer ego-motion dynamics from visual observations using \textit{EgoDyn-Bench}. Our experiments analyze (i) differences across model families, (ii) the impact of domain priors, and (iii) the influence of explicit dynamic information on motion understanding.

\subsection{Baselines}
We evaluate a set of non-foundation model baselines that estimate ego-motion directly from visual input, spanning both classical and learning-based approaches. Specifically, we consider (i) classical optical flow based on motion field formulations~\cite{HORN1981185,LonguetHiggins}, (ii) feature-based visual odometry using KLT tracking with essential matrix estimation~\cite{lucas1981iterative,323794,Hartley_Zisserman_2004}, (iii) learned optical flow using RAFT~\cite{teed2020raftrecurrentallpairsfield}, and (iv) learning-based visual odometry using TartanVO~\cite{wang2020tartanvogeneralizablelearningbasedvo}. The baselines share the same visual input and are evaluated against the same oracle-derived ground-truth labels as the foundation models. To produce 
answers, they apply heuristic mapping rules to their estimated motion signals, for example, optical flow magnitude to speed trend, using the same 
question categories and answer spaces defined by the oracle. Further details on each baseline are available in the supplementary material.

\subsection{Evaluated Model Families}
We evaluate representative models from three categories: (i) closed-source multimodal foundation models (\eg, GPT-5.1, Claude), (ii) open-source vision-language models (\eg, Qwen-VL), and (iii) domain-specific architectures for physically grounded reasoning (\eg, RoboTron-Drive). This categorization, detailed fully in Tables~\ref{tab:vision-only} and \ref{tab:vision-trajectory}, enables a direct comparison between general-purpose models and those with domain-specific inductive biases.

\subsection{Evaluation Protocol}

\textit{EgoDyn-Bench} comprises 14,000 QA pairs across 1,000 balanced 3-second driving scenarios (500 real-world, 500 simulated). We evaluate 14 question categories (direct and temporal dynamics) via deterministic binary and multiple-choice templates. Full parsing rules, prompts, and code are available in the supplementary material. As introduced in \Cref{subsec:metrics}, we report balanced accuracy, Macro-F1, temporal accuracy, and the introduced WPCR to assess both semantic correctness and physical coherence.

\subsection{Input Settings and Evaluation Axes}

To analyze the contribution of visual motion cues, we evaluate two input settings:
\textbf{(i) Vision-only:} Models receive only visual observations. We uniformly sample 10 frames from each 3-second clip ($\approx$ 3.3 FPS). This sampling rate preserves sufficient temporal resolution for macroscopic dynamic reasoning while maintaining computational feasibility across the extensive suite of evaluated models. 
\textbf{(ii) Vision $\boldsymbol{+}$ Dynamics:} Models additionally receive explicit ego-motion signals as structured text. To isolate optimal representation formats, we ablate \textit{four} textual trajectory encodings: a high-level \textit{Summary} (8 scalar statistics covering kinematic means and extrema, \eg, max/mean speed, max lateral acceleration), a dense kinematic \textit{Timeseries} (per-channel $v, a, \omega, j$ values at $N$ evenly-spaced timesteps), spatial \textit{Coordinates} (zero-centered $x, y$ waypoints and heading $\theta$ at $N$ timesteps), and a \textit{Full} combination of both timeseries and coordinate data. All explicitly provided kinematic information is temporally aligned with the $N$ subsampled images. 

Our subsequent analysis is structured along three complementary axes: (i) assessing vision-only performance to test implicit motion extraction, (ii) ablating textual dynamics to isolate reliance on numerical signals, and (iii) comparing visual domains to separate reasoning deficits from dataset artifacts.

\section{Results \& Discussion}
We evaluate the ego-motion reasoning capabilities of vision-centric foundation models across three complementary axes: \textit{(i) vision-only performance} to assess implicit motion extraction from visual input, \textit{(ii) dynamics-informed question answering} to quantify the contribution of explicit trajectory signals, and \textit{(iii) domain invariance} to separate reasoning deficits from dataset artifacts.

\subsection{Vision-Centric Study}\label{subsec:vision-only}
Our initial analysis investigates the models' ability to infer semantic ego-motion concepts directly from visual observations without auxiliary state information, probing their capacity for zero-shot physical grounding from visual cues alone. The results, summarized in \Cref{tab:vision-only}, reveal three key insights: 

\noindent\textbf{(i) Classical Baselines Outperform VLMs.} Vision-centric foundation models exhibit a significant performance gap compared to simple non-foundation model approaches. While classical baselines are restricted to the geometrically-answerable subset (6/14 questions), they outperform even the largest closed-source MLLMs on this overlapping subset (Visual Odometry baseline: BAcc 63.8\% vs. GPT-5.1: 55.1\%, Gemini 3 Pro: 59.6\%, Qwen3-VL-8B: 52.3\%), underscoring a fundamental struggle in current architectures to extract low-level kinematic representations from visual input. Temporally shuffling the input frames leaves performance unchanged (BAcc 39.0 vs. 38.9 ordered), indicating the deficit lies not in the visual encoder but in downstream temporal integration.

\noindent\textbf{(ii) Scaling and Domain Paradox.} Scaling to larger closed-source MLLMs yields only marginal gains over smaller open-source counterparts: the best closed-source model (Gemini 3, BAcc 47.0\%) outperforms the best open-source 8B model (Cosmos-Reason 2-8B, BAcc 39.9\%) by only 7.1\%, a negligible margin given the orders-of-magnitude difference in scale. Domain-specific VLAs perform comparably to or below general open-source models of equivalent size (RoboTronDrive: 38.6\%), indicating that neither scale nor in-domain training resolves the underlying architectural failure to ground physical reasoning in visual observation alone. Increasing temporal resolution to 10 FPS likewise yields no improvement (+0.4pp on BACC for Qwen3-VL, see Supplementary), confirming the bottleneck is structural rather than input-limited.

\noindent\textbf{(iii) Predictive Fallback Bias.} A consistent disparity between raw and balanced accuracy across VLMs indicates that models default to a single dominant answer when physical reasoning fails, rather than discriminating across classes. Our Visual Odometry baseline nearly eliminates this gap (65.1\% vs. 63.8\%), suggesting that explicit geometric representations effectively anchor predictions and mitigate response bias.

\begin{table}[!ht]
\centering
\begin{threeparttable} 
\caption{Metrics for vision-only ablation. All metrics are reported as percentages. Note that no explicit dynamic state inputs are provided for all models in this evaluation. Best values are presented \textbf{bold}; second-best are \underline{underlined.} $^1$ Evaluated on a functional subset of the questions. $^2$ Evaluated without visual observation ($O = \{\emptyset \}$). $^3$ Evaluated with static visual observation ($O = \{I_0\}$). $^4$ Evaluated with shuffled visual observations ($O^\prime = \left\{ I_{\sigma(i)} \right\}^N_{i=0}$), where the temporal order is randomized}
\label{tab:vision-only}
\scriptsize
\renewcommand{\arraystretch}{1.05}
\setlength{\tabcolsep}{2pt}

\begin{tabular}{@{} l c c c c c c c c @{}} 
\toprule

\multirow{2}{*}{\textbf{Model}} 
& \textbf{Parsable} 
& \multicolumn{3}{c}{\textbf{Semantic}} 
& \multicolumn{2}{c}{\textbf{Temporal}} 
& \multicolumn{2}{c}{\textbf{Consistency}} \\

\cmidrule(lr){2-2} \cmidrule(lr){3-5} \cmidrule(lr){6-7} \cmidrule(lr){8-9}

& \textbf{$\uparrow$} 
& \textbf{Acc. $\uparrow$} 
& \textbf{BAcc. $\uparrow$} 
& \textbf{F1 $\uparrow$} 
& \textbf{Acc. $\uparrow$} 
& \textbf{F1 $\uparrow$} 
& \textbf{WPCR $\uparrow$} 
& \textbf{PCov. $\uparrow$} \\

\midrule

\rowcolor{SectionGray}
\multicolumn{9}{c}{\textbf{Geometric Baselines}} \\ 
\midrule
\textit{Flow Heuristic}$^1$~\cite{HORN1981185,LonguetHiggins} & / & 58.4 & 47.0 & 42.2 & 48.5 & 43.9 & 46.5 & 79.6 \\
\textit{Visual Odometry}\tnote{1}~\cite{lucas1981iterative, 323794,Hartley_Zisserman_2004}  & / & 65.1 & \textbf{63.8} & \textbf{63.0} & \underline{62.7} & \underline{61.1} & 48.0 & 97.1 \\
\textit{RAFT Flow}\tnote{1}~\cite{teed2020raftrecurrentallpairsfield} & / & \textbf{69.8} & \underline{59.6} & \underline{56.8} & 53.9 & 58.2 & 47.5 & 85.3 \\
\textit{TartanVO}\tnote{1}~\cite{wang2020tartanvogeneralizablelearningbasedvo} & / & \underline{67.8} & 54.4 & 50.7 & \textbf{65.9} & \textbf{64.3} & 42.4 & \underline{99.8} \\

\midrule
\rowcolor{SectionGray}
\multicolumn{9}{c}{\textbf{Ablated Baselines}} \\ 
\midrule

\textit{Qwen3-VL-8B}\tnote{2}~\cite{bai2025qwen3vltechnicalreport} & 100.0 & 42.3 & 33.5 & 19.3 & 33.3 & 20.2 & 20.0 & \textbf{100.0} \\
\textit{Qwen3-VL-8B}\tnote{3}~\cite{bai2025qwen3vltechnicalreport} & 100.0 & 47.5 & 37.0 & 30.0 & 41.0 & 31.6 & 97.4 & 83.9 \\
\textit{Qwen3-VL-8B}\tnote{4}~\cite{bai2025qwen3vltechnicalreport} & 100.0 & 49.1 & 39.0 & 31.7 & 40.4 & 30.0 & \textbf{98.4} & 84.8 \\

\midrule

\rowcolor{SectionGray}
\multicolumn{9}{c}{\textbf{Closed-Source VLMs/MLLMs}} \\
\midrule
\textit{Claude Sonnet 4.5}~\cite{bai2022constitutionalaiharmlessnessai} & 99.7 & 49.6 & 38.0 & 32.9 & 36.8 & 34.9 & 83.4 & 94.4 \\
\textit{GPT-5.1}~\cite{openai2024gpt4technicalreport} & 100.0 & 54.3 & 45.2 & 40.1 & 43.0 & 38.7 & 96.3 & 76.9 \\
\textit{gemini-2.0-flash}~\cite{geminiteam2025geminifamilyhighlycapable} & 100.0 & 46.2& 36.5 & 32.4 & 37.2 & 35.1 & 54.1 & 97.5 \\
\textit{gemini-3-pro-preview}~\cite{geminiteam2025geminifamilyhighlycapable} & 94.3 & 53.6 & 47.0 & 44.3 & 47.3 & 45.2 & 59.3 & 98.4 \\

\midrule

\rowcolor{SectionGray}
\multicolumn{9}{c}{\textbf{Open-Source VLMs}} \\
\midrule
\textit{Qwen3-VL-2B}~\cite{bai2025qwen3vltechnicalreport} & 100.0 & 47.1 & 37.4 & 31.2 & 42.3 & 37.5 & 39.1 & 99.8 \\
\textit{Qwen3-VL-4B}~\cite{bai2025qwen3vltechnicalreport} & 100.0 & 49.7 & 39.8 & 34.4 & 39.7 & 35.1 & 82.9 & 90.0 \\
\textit{Qwen3-VL-8B}~\cite{bai2025qwen3vltechnicalreport} & 100.0 & 49.8 & 38.9 & 33.0 & 39.3 & 32.7 & \underline{97.9} & 84.6 \\
\textit{InternVL3-2B}~\cite{zhu2025internvl3exploringadvancedtraining} & 99.9 & 44.5 & 32.8 & 20.4 & 36.2 & 26.4 & 39.4 & 99.7 \\
\textit{InternVL3.5-4B}~\cite{wang2025internvl35advancingopensourcemultimodal} & 100.0 & 46.8 & 36.0 & 25.5 & 42.7 & 28.7 & 48.0 & 99.5 \\
\textit{InternVL3.5-8B}~\cite{wang2025internvl35advancingopensourcemultimodal}  & 100.0 & 47.0 & 38.8 & 30.0 & 42.7 & 33.5 & 48.6 & 78.0 \\
\textit{InternVL3.5-38B}~\cite{wang2025internvl35advancingopensourcemultimodal}  & 100.0 & 46.5 & 37.6 & 26.8 & 44.8 & 27.0 & 72.6 & 72.4 \\
\textit{Camreasoner-8B}~\cite{wu2026camreasonerreinforcingcameramovement} & 91.1 & 45.0 & 36.9 & 27.4 & 37.2 & 28.6 & 46.3 & 92.2 \\ 
\textit{Cosmos Reason 2-2B}~\cite{nvidia2025cosmosworldfoundationmodel} & 100.0 & 46.2& 35.1 & 23.6 & 39.5 & 25.8 & 72.7 & \textbf{100.0 }\\
\textit{Cosmos Reason 2-8B}~\cite{nvidia2025cosmosworldfoundationmodel} & 100.0 & 48.8 & 39.9 & 35.8 & 41.0 & 36.7 & 92.2 & 80.6 \\

\midrule

\rowcolor{SectionGray}
\multicolumn{9}{c}{\textbf{VLA Models}} \\
\midrule
\textit{ImpromptuVLA}~\cite{chi2025impromptuvlaopenweights} & 100.0 & 47.3& 37.8 & 28.9 & 40.2 & 29.2 & 41.1 & 72.8 \\
\textit{RoboTron-Drive}~\cite{huang2025robotrondriveallinonelargemultimodal} & 99.4 & 48.0 & 38.6 & 32.0 & 41.3 & 31.3 & 69.8 & 96.8 \\
\bottomrule
\end{tabular}
\end{threeparttable} 
\end{table}

\subsection{Dynamics-Informed Reasoning}
We evaluate the impact of providing explicit ego-motion signals as auxiliary input. The results in \Cref{tab:vision-trajectory} yield the following conclusions: 

\begin{table}[!ht]
\centering
\begin{threeparttable} 
\caption{Metrics for vision and trajectory ablation. All metrics are reported as percentages. Note that explicit dynamic state inputs are provided for all models in this evaluation. Further advanced embedding ablations are available in the supplementary material. Best values are in \textbf{bold}; second-best are \underline{underlined.} $^1$ Evaluated without visual observation ($O = \{\emptyset \}$).}
\label{tab:vision-trajectory}
\scriptsize
\renewcommand{\arraystretch}{1.05}
\setlength{\tabcolsep}{2pt}

\begin{tabular}{@{} l c c c c c c c c @{}} 
\toprule

\multirow{2}{*}{\textbf{Model}} 
& \textbf{Parsable} 
& \multicolumn{3}{c}{\textbf{Semantic}} 
& \multicolumn{2}{c}{\textbf{Temporal}} 
& \multicolumn{2}{c}{\textbf{Consistency}} \\

\cmidrule(lr){2-2} \cmidrule(lr){3-5} \cmidrule(lr){6-7} \cmidrule(lr){8-9}

& \textbf{$\uparrow$} 
& \textbf{Acc. $\uparrow$} 
& \textbf{BAcc. $\uparrow$} 
& \textbf{F1 $\uparrow$} 
& \textbf{Acc. $\uparrow$} 
& \textbf{F1 $\uparrow$} 
& \textbf{WPCR $\uparrow$} 
& \textbf{PCov. $\uparrow$} \\

\midrule

\rowcolor{SectionGray}
\multicolumn{9}{c}{\textbf{Baseline (Default: Summary Encoding)}} \\
\midrule
\textit{Qwen3-VL-8B}$^1$~\cite{bai2025qwen3vltechnicalreport} & 100.0& 65.9 &  59.6 & 54.1 & 53.1 & 46.3 & 33.0 & \textbf{100.0} \\
\textit{InternVL3.5-8B}~\cite{wang2025internvl35advancingopensourcemultimodal} & 100.0 & 61.9 & 55.7 & 49.4 & 55.3 & 45.4 & 17.5 & \textbf{100.0} \\

\midrule

\rowcolor{SectionGray}
\multicolumn{9}{c}{\textbf{Closed-Source VLMs/MLLMs (Default: Summary Encoding)}} \\
\midrule
\textit{Claude Sonnet 4.5}~\cite{bai2022constitutionalaiharmlessnessai} & 98.1& 71.5 & 63.5 & 61.1 & 58.7 & 58.1 & 97.1 & 91.2 \\
\textit{GPT-5.1}~\cite{openai2024gpt4technicalreport} & 100.0 & \underline{71.9} & \underline{66.1} & \underline{64.2} & 53.6 & 49.1 & 97.8 & 84.3 \\
\textit{gemini-2.0-flash}~\cite{geminiteam2025geminifamilyhighlycapable} & 100.0 & 65.1 & 55.8 & 52.6 & 59.4 & 57.1 & 66.6 & 98.9 \\
\textit{gemini-3-pro-preview}~\cite{geminiteam2025geminifamilyhighlycapable} & 98.8& \textbf{75.8} & \textbf{68.3} & \textbf{67.7} & 69.8 & 70.0 & 87.7 & 97.2 \\

\midrule

\rowcolor{SectionGray}
\multicolumn{9}{c}{\textbf{Open-Source VLMs (Default: Summary Encoding)}} \\
\midrule 
\textit{InternVL3-8B}~\cite{zhu2025internvl3exploringadvancedtraining} & 100.0 & 53.8 & 44.8 & 33.4 & 46.3 & 32.1 & 81.0 & 96.8 \\
\textit{Cosmos Reason 2-8B}~\cite{nvidia2025cosmosworldfoundationmodel} & 100.0 & 64.3& 56.5 & 54.0 & 52.8 & 51.1 & 71.4 & \underline{97.6} \\

\midrule

\rowcolor{SectionGray}
\multicolumn{9}{c}{\textbf{Open-Source VLA (Default: Summary Encoding)}} \\
\midrule 
\textit{ImpromptuVLA}~\cite{chi2025impromptuvlaopenweights} & 100.0& 55.6 & 52.1 & 45.6 & 48.0 & 39.8 & 39.7 & 90.3 \\
\midrule

\rowcolor{SectionGray}
\multicolumn{9}{c}{\textbf{Trajectory Encoding Ablation (Qwen3-VL-8B~\cite{bai2025qwen3vltechnicalreport})}} \\
\midrule
\textit{- Summary} & 100.0 & 63.6 & 54.7 & 52.7 & 43.4 & 41.1 & 89.7 & 90.9 \\
\textit{- Timeseries (Kinematics)} & 100.0 & 69.7 & 62.2 & 61.9 & \underline{76.7} & \underline{77.9} & 92.6 & 90.6 \\
\textit{- Coordinates (Spatial)} & 100.0 & 55.3 & 46.6 & 43.0 & 49.9 & 48.9 & \textbf{97.9} & 62.2 \\
\textit{- Full (Times. + Coord.)} & 100.0 & 69.0 & 61.3 & 60.4 & \textbf{77.3} & \textbf{78.4} & \underline{97.8} & 78.1 \\



\bottomrule
\end{tabular}
\end{threeparttable} 
\end{table}

\noindent\textbf{(i) Explicit Dynamics Consistently Improve Performance.} Integrating explicit dynamics improves performance across all models and metrics, confirming that the failures in \Cref{tab:vision-only} stem from a misalignment between visual observations and physical motion concepts, not from an absence of physical reasoning capacity. When provided with explicit kinematic data, models demonstrate the ability to reason about ego-motion that visual input alone fails to activate. While encoding identical physical information, trajectory-only inputs in different forms yield different balanced accuracy scores, indicating that models must process dynamic representations rather than simply forwarding subtle labels.

\noindent\textbf{(ii) Models Bypass Visual Input for Motion Reasoning.} The trajectory-informed experiments expose an asymmetry in modality importance. For Qwen3-VL-8B, replacing visual frames entirely with trajectory text yields a BAcc of $59.6\%$, a significant $+20.7$pp surge over the vision-only baseline ($38.9\%$). Reintroducing visual frames to this text-only baseline recovers a negligible $+2.6$pp gain under the best encoding strategy. Furthermore, with suboptimal encodings, performance regresses \textit{entirely} below the text-only baseline. This reveals a functional decoupling in current architectures: \textbf{ego-motion logic is derived almost exclusively from the language modality}, while visual observations serve as redundant or even interfering signals that the reasoning core fails to integrate. 

\noindent\textbf{(iii) Consistency Relies on Static Visual Context.} WPCR rises sharply from 20.0 with no visual input to 97.4 with a single static frame, but increases negligibly when additional frames are provided. This shows that physical consistency in model predictions is driven by the presence of any visual context, not by temporal reasoning over the frame sequence. Adding explicit trajectory text further improves WPCR, but reduces the contribution of visual input to near zero, confirming that motion reasoning is routed almost exclusively through the language modality.

\noindent\textbf{(iv) The Encoding Advantage.} Structured kinematic data consistently outperforms high-level semantic summaries across all models. With optimized encoding (Timeseries), smaller open-source models match or exceed closed-source MLLMs on temporal and consistency metrics, suggesting that representation quality is a more significant performance driver than parameter scale as is. Consequently, our conclusions suggest that simply scaling model size is insufficient for embodied tasks. Instead, future research must prioritize developing stronger physical alignment strategies during pre-training to bridge this vision-language gap.

\subsection{Isolating the Domain Gap}\label{subsec:domain_gap}
To verify that our results are driven by ego-motion complexity rather than a simulation-to-reality gap or style-transfer artifacts, we conduct a controlled ablation across three visual domains: (i) real-world data (\textit{nuScenes}), (ii) raw synthetic data, and (iii) style-transferred synthetic data used in these experiments.

As shown in \Cref{tab:source_comparison}, a representative VLM and baseline performance remain consistent across all three domains. This invariance indicates that the observed difficulties stem from a fundamental deficit in ego-motion reasoning rather than visual domain shifts. Notably, the stable performance of our geometric baselines across real and style-transferred sequences further supports the use of our synthetic data for assessing real-world ego-motion understanding.

\textbf{Natural Visual-Artifact Ablation.} 80 of 500 CARLA-transferred clips (16\%) contain spatial artifacts from upstream CARLA rendering. As these are temporally stable within each clip, optical flow is preserved while photometric quality is degraded. Per-clip accuracy on this subset differs from that of the other 420 by $\leq 3$ pp across leaderboard models, with mixed direction, further confirming the perception bottleneck: photometric quality is not meaningfully exploited. The subset list is released for downstream studies.

\begin{table}[!ht]
\centering
\caption{Comparison of balanced accuracy for  VLM and Baselines across different data sources. Best values are in \textbf{bold}; second-best are \underline{underlined.} $^1$ Representative VLM comparison point.}
\begin{threeparttable} 
\label{tab:source_comparison}
\scriptsize
\renewcommand{\arraystretch}{1.05}
\setlength{\tabcolsep}{6pt}

\begin{tabular}{@{} l c c c c @{}} 
\toprule

\textbf{Model} 
& \textbf{Real $\uparrow$} 
& \textbf{Sim $\uparrow$} 
& \textbf{Transf. $\uparrow$} 
& \textbf{$\Delta$ max $\downarrow$} \\

\midrule

\rowcolor{SectionGray}
\multicolumn{5}{c}{\textbf{Geometric Baselines}} \\
\midrule 
\textit{Flow Heuristic}~\cite{HORN1981185,LonguetHiggins} & 49.3 & 39.0 & 44.4 & 10.3 \\
\textit{Visual Odometry}~\cite{lucas1981iterative, 323794,Hartley_Zisserman_2004}  & \underline{63.5} & \textbf{62.4} & \textbf{64.0} & \underline{1.6} \\
\textit{RAFT Flow}~\cite{teed2020raftrecurrentallpairsfield} & \textbf{65.5} & \underline{60.5} & 54.6 & 10.9 \\
\textit{TartanVO}~\cite{wang2020tartanvogeneralizablelearningbasedvo} & 51.3 & 51.0 & \underline{56.0} & 5.0 \\

\midrule

\rowcolor{SectionGray}
\multicolumn{5}{c}{\textbf{VLM Baselines}} \\
\midrule
\textit{Qwen3-VL-8B}\tnote{1}~\cite{bai2025qwen3vltechnicalreport} & 40.0 & 39.7 & 41.0 & \textbf{1.3} \\

\bottomrule
\end{tabular}
\end{threeparttable} 
\end{table}

\section{Conclusion}
We introduced \textit{EgoDyn-Bench} to evaluate physical ego-motion understanding in vision-centric foundation models. Our audit of 20$+$ models reveals a consistent and severe \textbf{Perception Bottleneck}: despite possessing physically consistent internal reasoning, current VLMs and VLAs fail to ground it in visual observations, frequently lagging behind classical non-learned geometric baselines, a deficit that persists independently of model scale, domain-specific training, and visual domain. When explicit kinematic encodings are provided, performance recovers substantially across all models. This, however, exposes a \textbf{structural asymmetry}: ego-motion understanding is derived almost exclusively from the language modality, with visual observations contributing negligible temporal signal. This functional disentanglement between vision and language is the central architectural failure \textit{EgoDyn-Bench }diagnoses. Resolving it, through native alignment between dynamic representations and visual perception during pre-training, is the critical open challenge for physically grounded embodied AI.

\textbf{Limitations.}\textit{EgoDyn-Bench} targets ego-motion understanding over short horizons (3 s), where individual maneuvers are cleanly separable and attributable, extending to long-horizon and multi-agent scene-level reasoning is a natural next step. As a grounding diagnostic, it isolates whether models align physical concepts with visual observation rather than measuring closed-loop driving performance.

\textbf{Future Work.} To address the pure reliance on language modality to understand the motion kinematics, we aim to examine specific kinematic encoding paired with explicit alignment strategies for enhancement of vision-centric foundation models.



\bibliographystyle{splncs04}
\bibliography{references}
\appendix
\section*{Supplementary Content}
\label{sec:supplementary}

The Supplementary references additional clarifications, experiments, and ablations that were also referenced within the main paper. It is structured to provide a comprehensive overview of the dataset construction, extended experimental setups, robustness checks, and public assets.

\section{\textit{EgoDyn-Bench} Construction}

\subsection{Data Curation \& Balancing}
Real-world driving datasets are inherently affected by a long-tail distribution problem: the vast majority of driving logs consist of steady, straight-line motion, while critical dynamic events (e.g., emergency braking, high lateral acceleration, or evasive maneuvers) are exceedingly rare. Randomly sampling from such datasets yields an imbalanced benchmark that rewards models for simply predicting the most frequent, nominal driving state (a ``mode collapse'' in reasoning). 

To ensure \textit{EgoDyn-Bench} robustly evaluates the full spectrum of physical ego-motion, we combine real-world sequences from nuScenes with augmented sequences simulated in CARLA. We then apply a deterministic, multi-objective greedy selection algorithm to extract a perfectly balanced final benchmark of 1,000 clips.

\subsubsection{Data Curation Pipeline.}
\label{sec:data_curation_pipeline}
To transform raw driving logs into a rigorous evaluation set, we implemented a strict, multi-stage data curation pipeline encompassing extraction, kinematic smoothing, and rigorous quality assurance according to~\cite{gao2026stylevladrivingstyleawarevision}.

\textbf{Raw Clip Extraction.} 
We standardized all sequences to 3-second temporal windows sampled uniformly at 10 Hz. For real-world data (\textit{nuScenes}), we extracted 3-second backward-looking clips anchored at annotated keyframes, enforcing a minimum threshold of 20 valid camera frames per clip to ensure visual continuity. For simulated data, we extracted non-overlapping 3-second windows (stride of 30 frames) from continuous CARLA \textit{Frenetix} replay logs. 

\textbf{Kinematic Feature Extraction \& Smoothing.}
A major challenge in physical state estimation is the amplification of high-frequency sensor noise during derivation (e.g., calculating jerk from raw position). To mitigate this, we applied Savitzky-Golay smoothing to the raw ego-poses at every derivative stage. This allowed us to robustly extract instantaneous speed, longitudinal acceleration, yaw rate, and jerk. Summary statistics (minimum, maximum, mean, and specific percentiles) were subsequently computed and stored for each sequence to serve as the basis for our semantic thresholds.

\textbf{QA Generation \& Traceability.}
Using a customized registry pattern, we implemented 12 distinct labeling rule types (e.g., single-threshold, sequential event, and trend analysis). Applying the 14 question templates to our entire curated data pool yielded approximately 42,000 candidate Question-Answer (QA) pairs. Crucially, every generated QA record retains full traceability: it stores the specific rule invoked, the exact parameters applied, and the computed kinematic evidence used to arrive at the answer. 

\textbf{Stratification \& Quality Assurance.}
Before passing the data pool to our balancing algorithm, we enforced strict data validation. On the array level, we verified timestamp monotonicity, correct tensor shapes, and the absence of NaN/Inf values. On the QA level, we verified schema completeness and valid answer assignments. Finally, to aid in downstream balancing, each clip was assigned binary stratification tags (\texttt{has\_turn}, \texttt{has\_braking}, \texttt{has\_aggressive}), mapping the pool into 8 distinct kinematic bins to ensure diverse coverage prior to greedy selection.

\subsubsection{Greedy Balancing Algorithm.}
Let $\mathcal{Q}$ be the set of all categorical question types evaluated in the benchmark. For each question $q \in \mathcal{Q}$, let $\mathcal{C}_q$ represent its set of possible answer classes. Our objective is to select a subset of $N=1000$ clips that achieves an approximately uniform distribution across all answer classes for every question, subject to a strict source-ratio constraint (50\% nuScenes, 50\% CARLA).

The target frequency for any answer class $c$ in question $q$ is defined as $f^*_{q,c} = 1 / |\mathcal{C}_q|$. 
At each step of the selection process, we maintain the current empirical frequency $\hat{f}_{q,c}$ of each answer class within the currently selected subset. The algorithm proceeds iteratively until $N$ clips are selected:

\begin{enumerate}
    \item \textbf{Identify Maximum Imbalance:} We identify the question $q_{worst}$ that exhibits the maximum deviation from its uniform target distribution, and isolate its most underrepresented answer class $c_{worst}$:
    \begin{equation}
        q_{worst}, c_{worst} = \arg\max_{q \in \mathcal{Q}, c \in \mathcal{C}_q} (f^*_{q,c} - \hat{f}_{q,c})
    \end{equation}
    
    \item \textbf{Candidate Filtering:} We retrieve all unselected clips from the data pool that feature the answer $c_{worst}$ for question $q_{worst}$. We filter these candidates to enforce the dataset source caps (maximum 500 clips per source).
    
    \item \textbf{Secondary Multi-Question Optimization:} Because a single clip contains answers to all 14 questions, selecting a clip to balance $q_{worst}$ will inherently alter the distributions of all other questions. To optimize global balance, we compute a secondary helpfulness score $H_i$ for each candidate clip $i$. If clip $i$ has answer $a_q$ for question $q$, its score is the sum of the deficits it helps resolve across all questions:
    \begin{equation}
        H_i = \sum_{q \in \mathcal{Q} \setminus \{q_{worst}\}} \max(0, f^*_{q, a_q} - \hat{f}_{q, a_q})
    \end{equation}
    We select the candidate clip that maximizes $H_i$ and add it to the benchmark subset, updating all running frequencies. 
\end{enumerate}

A detailed pseudo code representation can be found in~\Cref{alg:greedy_balance}.

\begin{algorithm}[ht]
\caption{Multi-Objective Greedy Balancing Algorithm}\label{alg:greedy_balance}
\begin{algorithmic}[1]
\Require Data pool $\mathcal{P}$, target size $N$, target uniform distribution $f^*$, subset source caps $C_{src}$
\Ensure Balanced subset $\mathcal{S}$
\State $\mathcal{S} \gets \emptyset$
\State Initialize current empirical frequencies $\hat{f}_{q,c} \gets 0$ for all $q, c$
\While{$|\mathcal{S}| < N$}
    \State \textit{// 1. Identify the worst imbalance.}
    \State $q_{worst}, c_{worst} \gets \arg\max_{q, c} (f^*_{q,c} - \hat{f}_{q,c})$
    \State \textit{// 2. Candidate filtering}
    \State $\mathcal{V} \gets \{i \in \mathcal{P} \setminus \mathcal{S} \mid \text{clip } i \text{ answers } c_{worst} \text{ for } q_{worst}\}$
    \State Filter $\mathcal{V}$ to enforce source caps $C_{src}$
    \If{$\mathcal{V}$ is empty}
        \State $\mathcal{V} \gets \{i \in \mathcal{P} \setminus \mathcal{S} \mid \text{clip } i \text{ satisfies } C_{src}\}$ \Comment{Fallback}
    \EndIf
    \State \textit{// 3. Secondary multi-question optimization}
    \State $best\_score \gets -\infty$
    \State $best\_clip \gets \text{null}$
    \For{\textbf{each} candidate $i \in \mathcal{V}$}
        \State $H_i \gets \sum_{q \neq q_{worst}} \max(0, f^*_{q, a_q} - \hat{f}_{q, a_q})$
        \If{$H_i > best\_score$}
            \State $best\_score \gets H_i$
            \State $best\_clip \gets i$
        \EndIf
    \EndFor
    \State \textit{// 4. Update selection and frequencies}
    \State $\mathcal{S} \gets \mathcal{S} \cup \{best\_clip\}$
    \State Update $\hat{f}_{q,c}$ based on answers in $best\_clip$
\EndWhile
\State \Return $\mathcal{S}$
\end{algorithmic}
\end{algorithm}

\subsection{Semantic Abstraction \& Calibrated Thresholds}
To map continuous vehicle kinematics to discrete semantic concepts, the deterministic oracle utilizes a set of carefully calibrated thresholds. Where possible, these thresholds are grounded in domain standards (e.g., ISO and AASHTO guidelines) and compared to the references introduced in the main paper. For relative metrics (like braking intensity and jerk), thresholds were calibrated empirically against the dataset distribution to ensure meaningful class separation (targeting approximate percentiles: $P_{25}, P_{50}, P_{75}$).

The continuous signals are aggregated over the 3-second temporal window (via min, max, or mean operations) and evaluated against the following rules:

\begin{itemize}
    \item \textbf{Turn Direction:} Evaluated on the absolute maximum yaw rate. A deadzone of $\pm 0.04$ rad/s ($\sim 2.3^\circ$/s) filters out sensor noise and nominal lane-keeping drift. Values exceeding $0.04$ rad/s indicate an intentional left turn, and below $-0.04$ rad/s indicate a right turn.
    \item \textbf{Braking Intensity:} Evaluated on the minimum longitudinal acceleration. Categorized as \textit{emergency} ($< -1.59$ m/s$^2$), \textit{moderate} ($-1.59$ to $-0.89$ m/s$^2$), \textit{low} ($-0.89$ to $-0.18$ m/s$^2$), or \textit{none} ($> -0.18$ m/s$^2$).
    \item \textbf{Speed Regime:} Evaluated on maximum speed. Categorized as \textit{stopped} ($< 0.5$ m/s), \textit{slow} ($< 5.0$ m/s), \textit{urban} ($< 13.9$ m/s, i.e., 50 km/h), or \textit{highway} ($\ge 13.9$ m/s).
    \item \textbf{Driving Smoothness:} Evaluated on the mean absolute jerk. Categorized as \textit{smooth} ($\le 1.25$ m/s$^3$), \textit{moderate} ($1.25$ to $2.15$ m/s$^3$), or \textit{aggressive} ($> 2.15$ m/s$^3$).
    \item \textbf{Speed Trend:} Evaluated on mean acceleration. Following ISO 15622 (Adaptive Cruise Control) steady-state control error tolerances, a deadzone of $\pm 0.25$ m/s$^2$ is applied. Values outside this band imply intentional \textit{accelerating} or \textit{decelerating}.
    \item \textbf{High Lateral Acceleration:} Evaluated via peak $a_{lat} \approx v \cdot \omega$. Inspired by AASHTO ``Green Book'' comfort limits, values exceeding $2.0$ m/s$^2$ ($\sim 0.2g$) are flagged as \textit{yes}.
    \item \textbf{Significant Heading Change:} Flagged as \textit{yes} if the cumulative heading change exceeds $0.2618$ radians ($15^\circ$).
    \item \textbf{Extreme Maneuver:} A compound boolean rule flagged as \textit{yes} if maximum absolute jerk exceeds $20.0$ m/s$^3$ OR minimum acceleration drops below $-3.924$ m/s$^2$ (emergency braking limit).
    \item \textbf{Stop-and-Go:} Flagged as \textit{yes} if the vehicle transitions between a stopped state ($v < 0.5$ m/s) and a moving state ($v > 2.0$ m/s) within the clip.
    \item \textbf{Brake-then-Turn:} A temporal sequence rule requiring a valid braking event ($a < -1.5$ m/s$^2$) to be temporally followed by a turning event ($|\omega| > 0.1$ rad/s).
\end{itemize}

\subsubsection*{Cross-Platform Generalization.}
A critical consideration for autonomous driving benchmarks is whether the defined kinematic boundaries generalize across different vehicle platforms. To address this, our semantic abstraction strictly delineates between \textit{physics-anchored} values and \textit{percentile-calibrated} values. The physics-anchored thresholds represent absolute human comfort and safety limits, which generalize universally across standard passenger vehicle platforms. Conversely, the percentile-calibrated thresholds (e.g., braking intensity boundaries) are dataset-specific. To allow researchers to seamlessly adapt \textit{EgoDyn-Bench} to new vehicle platforms or specific Operational Design Domains (ODDs), we provide a dedicated calibration script (\texttt{calibrate\_thresholds.py}) within the codebase to automatically re-normalize these dataset-specific boundaries based on new target distributions.

\subsection{Labeling Rules \& The Deterministic Oracle}
By applying the thresholds defined above to the temporally aligned kinematic state vectors of the \textit{EgoDyn-Bench} dataset, the deterministic oracle automatically annotates all 1,000 video clips. This programmatic approach ensures zero human annotation bias and provides mathematically and physically grounded ground truth for model evaluation.

\subsection{Full Question Bank \& Answer Options}
The resulting benchmark comprises 14 distinct question templates spanning direct dynamics, comparative analysis, and temporal localization. The full question bank, along with the mutually exclusive answer choices for each template, is detailed in Table \ref{tab:question_bank}.

\begin{table}[ht]
\centering
\caption{The 14 question templates of \textit{EgoDyn-Bench}, their evaluation categories, and the mutually exclusive discrete answer choices.}
\label{tab:question_bank}
\resizebox{\textwidth}{!}{
\begin{tabular}{lp{7.5cm}l}
\toprule
\textbf{Category} & \textbf{Question Text} & \textbf{Answer Choices} \\
\midrule
\rowcolor{SectionGray}
\multicolumn{3}{c}{\textbf{Direct Dynamics}} \\

\midrule
Turn Direction & Is the vehicle turning left, right, or going straight? & [left, right, straight] \\
Braking Intensity & What is the intensity level of the vehicle's braking? & [emergency, moderate, low, none] \\
Speed Regime & What is the vehicle's speed regime? & [stopped, slow, urban, highway] \\
Driving Smoothness & How smooth is the driving based on jerk? & [smooth, moderate, aggressive] \\
Speed Trend & Is the vehicle accelerating, decelerating, or maintaining steady speed? & [accelerating, decelerating, steady] \\
Mean Speed & Is the mean speed below 5 m/s (18 km/h)? & [yes, no] \\
Heading Change & Does the vehicle change heading by more than 15 degrees? & [yes, no] \\
Extreme Maneuver & Does the vehicle perform an extreme maneuver (high jerk or hard braking)? & [yes, no] \\
Motion Axis & Is the vehicle's motion primarily longitudinal (speeding up/slowing down) or lateral (turning)? & [longitudinal, lateral, none] \\
Lateral Accel & Does the vehicle experience high lateral acceleration? & [yes, no] \\
Stop-and-Go & Does the vehicle exhibit stop-and-go behavior? & [yes, no] \\
Brake-Then-Turn & Does the vehicle brake and then turn (sequential maneuver)? & [yes, no] \\
\midrule
\rowcolor{SectionGray}
\multicolumn{3}{c}{\textbf{Comparative \& Temporal}} \\

\midrule
Speed Peak Half & Does the maximum speed occur in the first or second half of the sequence? & [first\_half, second\_half, no\_peak] \\
Contrastive Seq. & Comparing the first and second halves of the sequence, which half has more dynamic driving? & [first\_half, second\_half, similar] \\
\bottomrule
\end{tabular}}
\end{table}

\section{Extended Experimental Setup \& Baselines}

\subsection{Detailed Evaluation Protocol}
To ensure reproducible and fair comparisons across highly diverse foundation models, all predictions are graded using a standardized deterministic evaluation script. 

\subsubsection*{Robustness of Deterministic Parsing.}
To address potential concerns regarding formatting penalties, we analyzed the parsability of model outputs across our benchmark. Because models are instructed in the system prompt to answer with only the chosen option, the vast majority of responses naturally conform to the expected label space. In all reported results, unparsed responses are treated as incorrect predictions, ensuring that parsability issues penalize rather than artificially inflate model scores.

To handle minor deviations, paraphrases, and verbose reasoning, we employ a 4-stage deterministic parsing cascade:
\begin{enumerate}
    \item \textbf{Exact Match:} Direct alignment with the target label space.
    \item \textbf{Underscore Normalization:} Standardizing whitespace and punctuation (e.g., ``first half'' $\leftrightarrow$ ``first\_half'').
    \item \textbf{Last-Line Extraction:} Isolating the final conclusion from chain-of-thought or verbose outputs.
    \item \textbf{Word-Boundary Substring Match:} Extracting the target label if it is unambiguously embedded within the final statement (e.g., ``The answer is: yes'' $\rightarrow$ ``yes'').
\end{enumerate}

\textbf{Parsability Rates and Metric Impact.} Table \ref{tab:parsability} details the parsing success rates for representative models out of $N=14,000$ total predictions. The $\Delta$ BAcc column represents the maximum possible inflation on the Balanced Accuracy metric if unparsed answers were excluded rather than penalized.

\begin{table}[ht]
\centering
\caption{Parsability rates and their maximum impact on Balanced Accuracy (BAcc). The $\Delta$ BAcc column represents the score difference between evaluating only parsed answers versus penalizing unparsed answers as incorrect. Where $N$ is the number of predictions.}
\label{tab:parsability}
\begin{tabular}{lccc}
\toprule
\textbf{Model Type} & \textbf{$N$} & \textbf{Parse Rate (\%)} & \textbf{$\Delta$ BAcc} \\
\midrule
Open-weight (Qwen3, InternVL, etc.) & 14,000 & 99.9 -- 100.0 & +0.0 \\
GPT-5.1 & 14,000 & 100.0 & +0.0 \\
Claude Sonnet 4.5 & 14,000 & 99.7 & +0.0 \\
Gemini 3 Pro (vision-only) & 14,000 & 94.3 & +1.5 \\
Gemini 3 Pro (w/ summary trajectory) & 14,000 & 98.8 & $<+0.5$ \\
CamReasoner & 14,000 & 91.1 & +3.2 \\
\bottomrule
\end{tabular}
\end{table}

The maximum observed inflation is 3.2 percentage points (CamReasoner), while the vast majority of evaluated models exhibit zero metric inflation. This confirms that ranking and performance trends discussed in the main paper are driven by genuine physical reasoning capabilities, not parsing artifacts.

\textbf{Taxonomy of Failure Modes.} An analysis of the unparsed responses reveals that failures are almost exclusively cases where models refuse to commit to an answer, which no deterministic parser could faithfully recover. We identify three primary failure modes: verbose reasoning without a conclusion, truncated responses (hitting the token limit mid-sentence), and exceedingly rare (<0.1\%) empty responses. For open-weight models, deterministic decoding (temperature set to $0.0$) yields near-perfect label adherence, rendering more complex constrained decoding techniques (e.g., grammar-based sampling via vLLM) unnecessary.

\subsection{Further Baseline Information}
To establish a lower bound for physical ego-motion understanding, we evaluate non-foundation model baselines that estimate dynamics directly from visual input. Because these classical methods lack semantic reasoning capabilities, we design explicit heuristic mapping rules to translate their continuous state outputs into the discrete semantic space of \textit{EgoDyn-Bench}.

\subsubsection{B.2.1 Optical Flow Baseline.}
Our first baseline utilizes dense optical flow to derive pixel-domain proxy signals for ego-motion. We use Farneback’s algorithm to compute the dense flow field between consecutive frames.

\textbf{ Preprocessing and Region of Interest.} To ensure computational stability and filter out irrelevant environmental noise, frames are converted to grayscale, down-sampled to a maximum width of 320 pixels, and smoothed via a Gaussian blur. We restrict all flow aggregation to a central horizontal band to filter irrelevant parts of the scene.

\textbf{Kinematic Proxy Signals.} Let ($f_x, f_y$) represent the optical flow vector at a pixel location ($x, y$). We define the image center as ($c_x, c_y$) and compute the relative pixel offsets $\Delta x = x-c_x$ and $\Delta y = y - c_y$, with the radial distance $r= \sqrt{(\Delta x)^2 + (\Delta y)^2}$. We compute three unitless, median-aggregated proxy signals per frame pair: 
\begin{enumerate}
    \item \textbf{Turn Score:} A proxy for rotational motion, derived from the tangential flow component. Positive values indicate counter-clockwise rotation (a left turn). 
    \begin{equation}
        S_{turn} = \text{median}\left( \frac{f_x\Delta y - f_y \Delta x}{r}\right)
    \end{equation}
    \item \textbf{Expansion Score:} A proxy for longitudinal acceleration, derived from the radial flow component. Positive values indicate outward radial expansion (accelerating).
    \begin{equation}
        S_{exp} = \text{median}\left( \frac{f_x \Delta x - f_y \Delta y}{r}\right)
    \end{equation}
    \item \textbf{Motion Magnitude:} A proxy for overall scene displacement.
    \begin{equation}
        M_{Mag} = \text{median} \left( \sqrt{f_x^2 + f_y^2}\right)
    \end{equation}
\end{enumerate}

\textbf{Heuristic Semantic Mapping.} Because monocular optical flow cannot reliably resolve absolute metric scale, this baseline is restricted to the subset of 6 questions that can be answered via qualitative motion patterns. We define a set of calibrated heuristic thresholds: $\tau_{turn} = 0.05$, $\tau_{exp} = 0.2$, $\tau_{lat} = 1.5$, $\tau_{head} = 3.0$, $\tau_{stop} = 0.3$, and $\tau_{move} = 1.5$. 

Using the temporally aggregated signals over the 3-second window, we map the continuous proxies to the discrete \textit{EgoDyn-Bench} semantic space $\mathcal{R}$ as follows:

\begin{enumerate}
    \item \textbf{Turn Direction:} Evaluated via the mean tangential flow proxy ($\bar{S}_{turn}$):
    \begin{equation}
    R_{turn} = 
    \begin{cases} 
    \text{left}, & \text{if } \bar{S}_{turn} > \tau_{turn} \\
    \text{right}, & \text{if } \bar{S}_{turn} < -\tau_{turn} \\
    \text{straight}, & \text{otherwise}
    \end{cases}
    \end{equation}
    
    \item \textbf{Speed Trend:} Evaluated via the mean radial expansion proxy ($\bar{S}_{exp}$):
    \begin{equation}
    R_{speed} = 
    \begin{cases} 
    \text{accelerating}, & \text{if } \bar{S}_{exp} > \tau_{exp} \\
    \text{decelerating}, & \text{if } \bar{S}_{exp} < -\tau_{exp} \\
    \text{steady}, & \text{otherwise}
    \end{cases}
    \end{equation}

    \item \textbf{High Lateral Acceleration:}
    \begin{equation}
    R_{lat} = 
    \begin{cases} 
    \text{yes}, & \text{if } \max(|S_{turn}|) > \tau_{lat} \\
    \text{no}, & \text{otherwise}
    \end{cases}
    \end{equation}

    \item \textbf{Significant Heading Change:}
    \begin{equation}
    R_{head} = 
    \begin{cases} 
    \text{yes}, & \text{if } \sum |S_{turn}| > \tau_{head} \\
    \text{no}, & \text{otherwise}
    \end{cases}
    \end{equation}

    \item \textbf{Stop-and-Go:} Requires detecting a temporal sequence where the motion magnitude $M_{mag}^{(t)}$ at time step $t$ transitions from a stopped state to a moving state:
    \begin{equation}
    R_{stop\_go} = 
    \begin{cases} 
    \text{yes}, & \begin{aligned}[t]
                  &\text{if } \exists\ t_1, t_2 \text{ such that } t_1 < t_2, \\
                  &M_{mag}^{(t_1)} < \tau_{stop} \text{ and } M_{mag}^{(t_2)} > \tau_{move}
                  \end{aligned} \\
    \text{no}, & \text{otherwise}
    \end{cases}
    \end{equation}

    \item \textbf{Brake-then-Turn:} Requires detecting a compound maneuver where a braking proxy is temporally followed by a turning proxy:
    \begin{equation}
    R_{brake\_turn} = 
    \begin{cases} 
    \text{yes}, & \begin{aligned}[t]
                  &\text{if } \exists\ t_1, t_2 \text{ such that } t_1 < t_2, \\
                  &S_{exp}^{(t_1)} < -\tau_{exp} \text{ and } |S_{turn}^{(t_2)}| > \tau_{turn}
                  \end{aligned} \\
    \text{no}, & \text{otherwise}
    \end{cases}
    \end{equation}
\end{enumerate}

\subsubsection{B.2.2 Visual Odometry Baseline.}
Our second geometric baseline is a proxy for visual odometry. Unlike full SLAM systems, this baseline is not designed to recover absolute scale or a full 6-DoF pose. Instead, it utilizes sparse feature tracking and essential matrix decomposition to estimate unitless per-frame-pair ego-rotation and translational magnitude proxies.

\textbf{Feature Tracking and Preprocessing.}
We convert frames to grayscale and apply a binary region-of-interest mask to isolate the central 60\% of the image, filtering out featureless sky and specular reflections from the ego-vehicle's hood. We detect up to 800 Shi-Tomasi corners and track them across frame pairs using the Pyramidal Lucas-Kanade (KLT) algorithm~\cite{lucas1981iterative}. Erroneous tracks with a pixel displacement exceeding 50 pixels are discarded. 

\textbf{Kinematic Proxy Signals.}
Let $\Delta p$ represent the pixel displacement of valid tracks between two consecutive frames. We compute two primary proxy signals:
\begin{enumerate}
    \item \textbf{Translational Proxy ($M_{disp}$):} Evaluated as the median displacement magnitude of all valid tracked features: $M_{disp} = \text{median}(||\Delta p||_2)$.
    \item \textbf{Rotational Proxy ($\theta$):} Assuming a default pinhole camera model, we robustly estimate the essential matrix $E$ using RANSAC. We decompose $E$ to recover the rotation matrix $R$, from which we extract the yaw angle $\theta = \arctan(R_{0,2} / R_{2,2})$. Positive values indicate a left turn.
\end{enumerate}

To ensure numerical stability, if the translational proxy is near-zero ($M_{disp} < 0.3$), the essential matrix decomposition becomes degenerate, and we enforce $\theta = 0^\circ$. If RANSAC yields fewer than 15 inliers, we fall back to a horizontal flow heuristic, approximating yaw via the median horizontal track displacement.

\textbf{Heuristic Semantic Mapping}
We apply the temporally aggregated signals over the 3-second window to the following calibrated thresholds: $\tau_{yaw} = 0.03^\circ$, $\tau_{peak} = 0.15^\circ$, $\tau_{stop} = 0.5$, $\tau_{move} = 2.0$, $\tau_{trend} = 0.3$, $\tau_{head} = 1.5^\circ$, $\tau_{lat} = 0.8^\circ$, and a fractional braking drop $\tau_{brake} = 0.4$. We map these continuous proxies to the semantic space $\mathcal{R}$ as follows:

\begin{align}
    \intertext{1. \textbf{Turn Direction:} Evaluated via the mean yaw ($\bar{\theta}$) and peak absolute yaw ($\theta_{peak} = \max(|\theta|)$):}
    R_{turn} &= 
    \begin{cases} 
    \text{left}, & \text{if } \bar{\theta} > \tau_{yaw} \text{ and } \theta_{peak} > \tau_{peak} \\
    \text{right}, & \text{if } \bar{\theta} < -\tau_{yaw} \text{ and } \theta_{peak} > \tau_{peak} \\
    \text{straight}, & \text{otherwise}
    \end{cases} \\
    \intertext{2. \textbf{Speed Trend:} Evaluated via the linear slope $m_{disp}$ of the displacement magnitude $M_{disp}$ over time:}
    R_{speed} &= 
    \begin{cases} 
    \text{accelerating}, & \text{if } m_{disp} > \tau_{trend} \\
    \text{decelerating}, & \text{if } m_{disp} < -\tau_{trend} \\
    \text{steady}, & \text{otherwise}
    \end{cases} \\
    \intertext{3. \textbf{High Lateral Acceleration:}}
    R_{lat} &= 
    \begin{cases} 
    \text{yes}, & \text{if } \theta_{peak} > \tau_{lat} \\
    \text{no}, & \text{otherwise}
    \end{cases} \\
    \intertext{4. \textbf{Significant Heading Change:}}
    R_{head} &= 
    \begin{cases} 
    \text{yes}, & \text{if } \sum |\theta| > \tau_{head} \\
    \text{no}, & \text{otherwise}
    \end{cases} \\
    \intertext{5. \textbf{Stop-and-Go:} Requires detecting a temporal sequence where the displacement magnitude $M_{disp}^{(t)}$ at time step $t$ transitions from a stopped state to a moving state:}
    R_{stop\_go} &= 
    \begin{cases} 
    \text{yes}, & \begin{aligned}[t]
                  &\text{if } \exists\ t_1, t_2 \text{ such that } t_1 < t_2, \\
                  &M_{disp}^{(t_1)} < \tau_{stop} \text{ and } M_{disp}^{(t_2)} > \tau_{move}
                  \end{aligned} \\
    \text{no}, & \text{otherwise}
    \end{cases} \\
    \intertext{6. \textbf{Brake-then-Turn:} Let the dynamic braking threshold be $\Delta_{brake} = \tau_{brake} \cdot \bar{M}_{disp}$. This requires detecting a sequence where a sharp drop in displacement is followed by a significant yaw:}
    R_{brake\_turn} &= 
    \begin{cases} 
    \text{yes}, & \begin{aligned}[t]
                  &\text{if } \exists\ t_1, t_2 \text{ such that } t_1 < t_2, \bar{M}_{disp} > 0.5,\\
                  &M_{disp}^{(t_1)} < (M_{disp}^{(t_1-1)} - \Delta_{brake}) \text{ and } |\theta^{(t_2)}| > \tau_{yaw}
                  \end{aligned} \\
    \text{no}, & \text{otherwise}
    \end{cases}
\end{align}

\subsubsection*{B.2.3 Learned Optical Flow Baseline (RAFT)}

To isolate whether the limitations of the classical flow heuristic stem from the rigid semantic mapping or the inadequacy of classical motion field estimation, we implement a learned optical flow alternative.

\textbf{Architecture and Weights.}
We replace the classical optical flow algorithm with the state-of-the-art RAFT (Recurrent All-Pairs Field Transforms) architecture. We use the \texttt{raft\_large} model, which benefits from extensive pre-training across a diverse set of datasets.

\textbf{Preprocessing and Signal Extraction.}
Visual inputs are converted to RGB tensors, downsampled to a maximum width of 320 pixels, and padded to ensure spatial dimensions are multiples of 8, a structural requirement of the RAFT network. After passing the frame pairs through the model, we extract the final high-resolution flow field from the refinement iterations. We remove the padding and apply the exact same region-of-interest cropping as defined in the classical baseline.

\textbf{Heuristic Semantic Mapping.}
To ensure a strictly controlled comparison between classical and learned perception backends, the physical proxy extraction and semantic mapping remain strictly identical to the classical flow baseline. We apply the same continuous radial and tangential decomposition to the RAFT flow vectors to extract the Turn Score ($S_{turn}$), Expansion Score ($S_{exp}$), and Motion Magnitude ($M_{mag}$), and apply the identical thresholding logic and sequential rules defined in Section B.2.1.

\subsubsection{B.2.4 Learned Visual Odometry Baseline.}

To evaluate whether the limitations of the visual odometry proxy stem from the classical KLT feature tracking pipeline, we implement a learning-based monocular VO alternative. We use TartanVO, a model trained on diverse synthetic scenes (TartanAir) that generalizes to real-world driving environments without fine-tuning.

\textbf{Preprocessing and Architecture}
Visual inputs are converted to RGB tensors and scaled/center-cropped to $640 \times 448$, matching the native resolution of the TartanVO network. The frame pairs are passed through the model alongside a scaled intrinsic matrix assuming default TartanAir parameters ($f_x = f_y = 320.0$, $c_x = 320.0$, $c_y = 240.0$). 

\textbf{Signal Extraction}
The network outputs a normalized 6-DoF relative pose vector for each frame pair. After denormalizing the outputs using the dataset-specific pose standard deviations, we extract the translational and rotational proxies:
\begin{enumerate}
    \item \textbf{Translational Proxy ($M_{disp}$):} Computed as the Euclidean norm of the predicted translation vector $\mathbf{t} = [t_x, t_y, t_z]^T$: 
    \begin{equation}
    M_{disp} = ||\mathbf{t}||_2
    \end{equation}
    \item \textbf{Rotational Proxy ($\theta$):} Extracted from the yaw component ($r_z$) of the predicted rotation vector and converted from radians to degrees.
\end{enumerate}

\textbf{Heuristic Semantic Mapping}
To maintain a controlled evaluation, we retain the exact same heuristic mapping logic and temporal sequence constraints defined for the classical VO baseline in Section B.2.2. However, because TartanVO's displacement magnitude and yaw outputs operate on a different scale space than pixel-domain KLT tracking, we recalibrate the empirical decision thresholds: $\tau_{yaw} = 0.5^\circ$, $\tau_{peak} = 1.0^\circ$, $\tau_{stop} = 0.15$, $\tau_{move} = 0.5$, $\tau_{trend} = 0.05$, $\tau_{head} = 5.0^\circ$, $\tau_{lat} = 2.0^\circ$, and $\tau_{brake} = 0.3$. 

By swapping only the perception backend while holding the reasoning logic constant, we confirm that the reasoning bottleneck persists even with state-of-the-art deep feature representations.

\section{Additional Analysis \& Robustness}

\subsection{Sensitivity Analysis}

A fundamental component of \textit{EgoDyn-Bench} is the deterministic oracle, which relies on calibrated kinematic thresholds to map continuous vehicle states to discrete semantic concepts. A critical methodological question is whether the evaluation results and the relative rankings of the evaluated foundation models are sensitive to the exact calibration of these thresholds.

To verify the robustness of our findings, we conduct a comprehensive sensitivity analysis. We uniformly perturb all numerical thresholds used by the oracle by a scalar factor $\alpha \in [0.5, 1.5]$. For each perturbation level, we regenerate the entire ground-truth label set and re-evaluate all models. 

\textbf{Model Ranking Stability.}
To quantify the stability of model performance across perturbation levels, we compute Kendall's rank correlation coefficient ($\tau$) between the model rankings at the nominal threshold ($\alpha = 1.0$) and the rankings at the perturbed thresholds. 

\begin{figure}[ht]
    \centering
    \includegraphics[width=\linewidth]{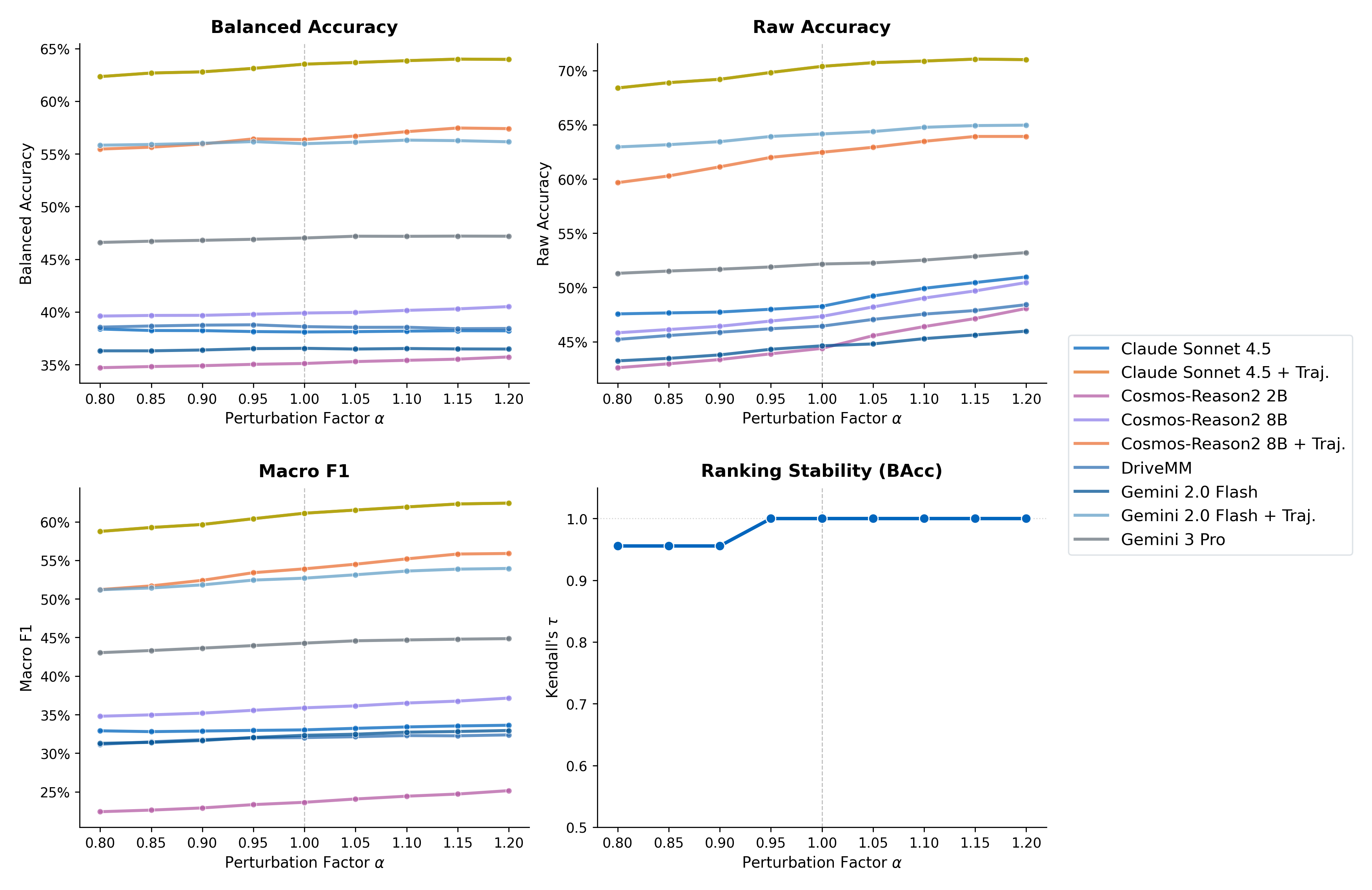}
    \caption{Global performance and ranking stability under threshold perturbation ($\alpha \in [0.5, 1.5]$). While raw and balanced accuracy exhibit minor scaling effects, Kendall's $\tau$ demonstrates that the relative ranking of models remains highly stable ($\tau > 0.9$) across almost all perturbation levels. This confirms that the observed perception bottleneck is robust to the specific kinematic calibration.}
    \label{fig:tau}
\end{figure}

As shown in~\Cref{fig:tau}, Kendall's $\tau$ remains above $0.90$ across the vast majority of question types, even under extreme threshold scaling ($\pm 50\%$). This exceptionally high correlation confirms that while absolute accuracy scores may shift slightly depending on the strictness of the maneuver definitions, the relative ordering of the models remains practically invariant. The observed ``Perception Bottleneck'' is therefore a structural property of the models, not an artifact of threshold selection.

\textbf{Consistency Metric Stability.}
Furthermore, we evaluate the stability of our physical consistency metrics. We track the behavior of the Weighted Physics Consistency Rate (WPCR) under the same threshold perturbations.

\begin{figure}[ht]
    \centering
    \includegraphics[width=0.75\linewidth]{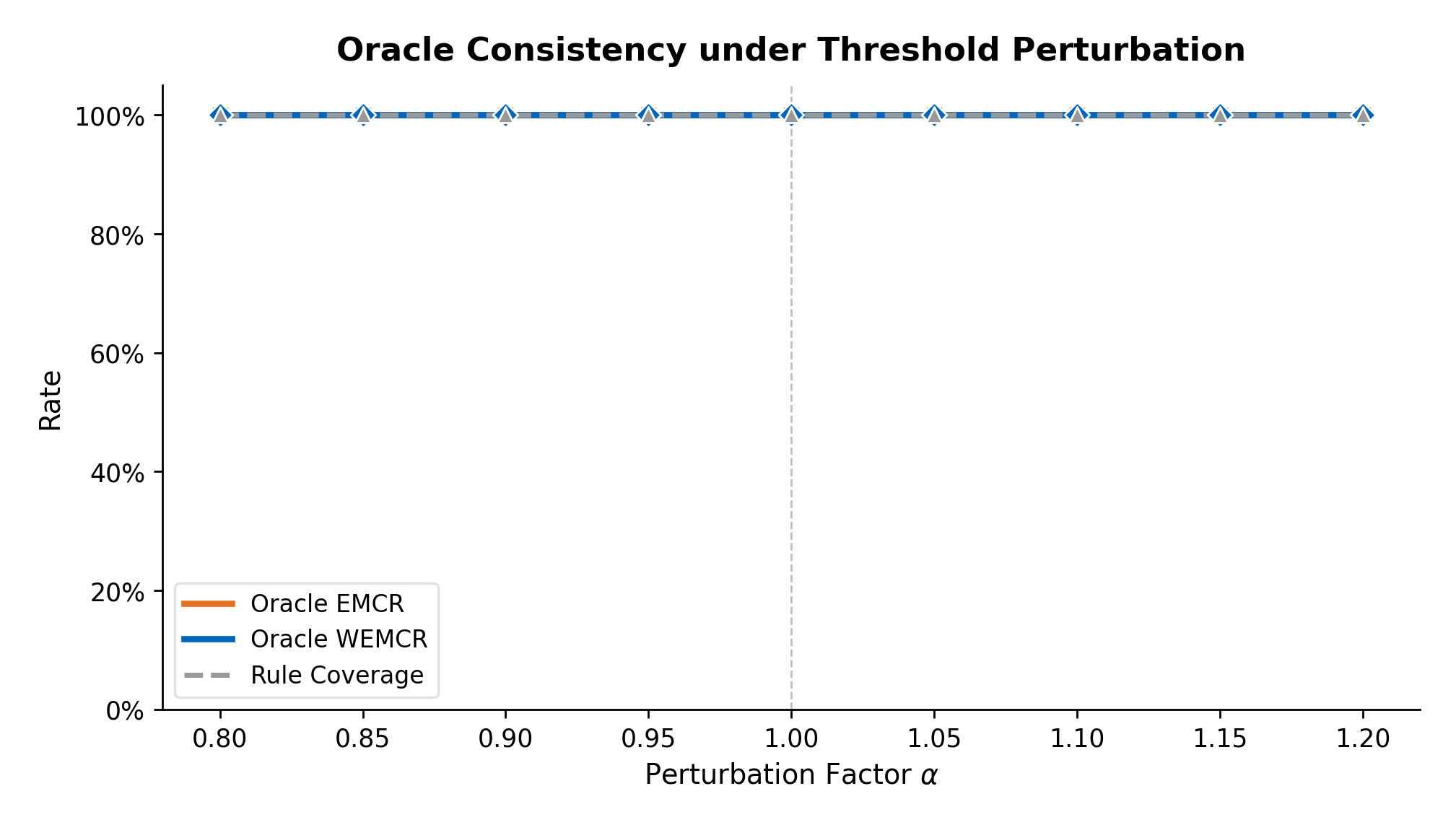}
    \caption{Stability of the deterministic oracle's physics-grounded consistency rules. The Weighted Physics Consistency Rate (WPCR) remains stable across the perturbation sweep, indicating that the Boolean implication logic is invariant to the specific scalar boundaries defining the maneuvers.}
    \label{fig:alpha}
\end{figure}

As shown in~\Cref{fig:alpha}, the global consistency metrics remain stable across the entire sweep of $\alpha \in [0.5, 1.5]$. This indicates that the Boolean implication rules defining the physics of motion are intrinsically robust. The relationships between continuous dynamics (e.g., speed regimes vs. stop-and-go behavior) hold true regardless of the specific scalar boundary used to define the categories. 

Consequently, users of \textit{EgoDyn-Bench} can confidently adjust these thresholds to suit specific operational design domains (ODDs) or research requirements without invalidating the comparative benchmarking framework.

\subsection{Advanced Embedding Ablations}

In Section 4.4 of the main paper, we introduced the \textit{Vision + Dynamics} evaluation setting and demonstrated that explicitly providing kinematic states as text substantially improves model performance. To determine the optimal representation for these physical states, we ablated four distinct textual trajectory encodings. Here, we detail the exact structure of these embeddings. 

For a 3-second clip sampled at $N=10$ timesteps, the textual context is prepended to the standard vision prompt. The four embedding modes are defined as follows:

\textbf{1. Summary (Default Baseline):}
Provides 8 global scalar statistics extracted over the full temporal window. This includes kinematic extremes and aggregates: maximum and mean speed, minimum acceleration, maximum yaw rate, maximum and mean jerk, maximum lateral acceleration, and total heading change.
\begin{quote}
\textit{Example Prompt Text:} ``Vehicle dynamics: max\_speed = 8.2 m/s (30km/h), mean\_speed = 7.4 m/s, min\_accel = -1.23 m/s², max\_yaw\_rate = 0.042 rad/s, max\_jerk = 2.85 m/s³, mean\_jerk = 0.91 m/s³, max\_lat\_accel = 0.34 m/s², heading\_change = 0.126 rad.''
\end{quote}

\textbf{2. Timeseries (Kinematics):}
Provides a dense temporal sequence of raw dynamic channels (speed $v$, acceleration $a$, yaw rate $\omega$, and jerk $j$) aligned to the $N$ sampled image frames.
\begin{quote}
\textit{Example Prompt Text:} ``Vehicle dynamics (10 time-steps over 3.0s): \\
t(s): 0.00, 0.33, 0.67, 1.00, ... \\
speed (m/s): 7.1, 7.4, 7.8, 8.0, ... \\
accel (m/s²): 0.82, 0.65, 0.31, 0.05, ... \\
yaw\_rate (rad/s): 0.012, 0.018, 0.025, ...''
\end{quote}

\textbf{3. Coordinates (Spatial):}
Provides purely spatial tracking information via zero-centered $(x, y)$ waypoints and heading $\theta$, requiring the model to internally differentiate these positions to infer dynamics.
\begin{quote}
\textit{Example Prompt Text:} ``Vehicle trajectory (10 waypoints over 3.0s, metres): \\
t(s): 0.00, 0.33, 0.67, 1.00, ... \\
x(m): 0.0, 2.4, 4.9, 7.3, ... \\
y(m): 0.0, 0.1, 0.3, 0.5, ... \\
heading (rad): 1.571, 1.578, 1.589, ...''
\end{quote}

\textbf{4. Full (Timeseries + Coordinates):}
The union of both the Timeseries and Coordinates prompts provides both explicit dynamic derivatives and spatial positioning.

\subsubsection{Extended Experiments: Cross-Architecture Consistency}
While the main paper details the embedding ablation for the Qwen3-VL-8B architecture, a critical question is whether the observed representational preferences are architecture-agnostic. To investigate this, we extend the ablation to the InternVL model family, specifically evaluating InternVL3.5-8B across all four text modalities. 

\begin{table}[ht]
\centering
\begin{threeparttable}
\caption{Trajectory Encoding Ablation across architectures. Presenting both Qwen3-VL-8B (from the main paper) and InternVL3.5-8B demonstrates that the representational preference for explicit kinematic timeseries is consistent across different foundation model families.}
\label{tab:internvl_ablation}

\scriptsize
\renewcommand{\arraystretch}{1.05}
\setlength{\tabcolsep}{2pt}

\begin{tabular}{@{} l c c c c c c c c @{}} 
\toprule

\multirow{2}{*}{\textbf{Model}} 
& \textbf{Parsable} 
& \multicolumn{3}{c}{\textbf{Semantic}} 
& \multicolumn{2}{c}{\textbf{Temporal}} 
& \multicolumn{2}{c}{\textbf{Consistency}} \\

\cmidrule(lr){2-2} \cmidrule(lr){3-5} \cmidrule(lr){6-7} \cmidrule(lr){8-9}

& \textbf{$\uparrow$} 
& \textbf{Acc. $\uparrow$} 
& \textbf{BAcc. $\uparrow$} 
& \textbf{F1 $\uparrow$} 
& \textbf{Acc. $\uparrow$} 
& \textbf{F1 $\uparrow$} 
& \textbf{WPCR $\uparrow$} 
& \textbf{PCov. $\uparrow$} \\

\midrule

\rowcolor{SectionGray}
\multicolumn{9}{c}{\textbf{Trajectory Encoding Ablation (Qwen3-VL-8B~\cite{bai2025qwen3vltechnicalreport})}} \\
\midrule
\textit{- Summary} & \textbf{100.0 }& 63.6 & 54.7 & 52.7 & 43.4 & 41.1 & 89.7 & 90.9 \\
\textit{- Timeseries (Kinematics)} & \textbf{100.0} & \textbf{69.7} & \textbf{62.2} & \textbf{61.9} & \underline{76.7} & \underline{77.9} & 92.6 & \textbf{90.6} \\
\textit{- Coordinates (Spatial)} & \textbf{100.0} & 55.3 & 46.6 & 43.0 & 49.9 & 48.9 & \textbf{97.9} & 62.2 \\
\textit{- Full (Times. + Coord.)} & \textbf{100.0} & \underline{69.0} & \underline{61.3} & \underline{60.4 }& \textbf{77.3} & \textbf{78.4} & \underline{97.8} & 78.1 \\

\midrule
\rowcolor{SectionGray}
\multicolumn{9}{c}{\textbf{Trajectory Encoding Ablation (InternVL3.5-8B~\cite{wang2025internvl35advancingopensourcemultimodal} )}} \\
\midrule
\textit{- Summary} & \textbf{100.0} & 62.5 & 53.8 & 49.2 & 47.3 & 37.6 & 56.1 & \textbf{93.1} \\
\textit{- Timeseries (Kinematics)} & 97.4 & 65.6 & 57.4 & 55.9 & 65.1 & 65.2 & 95.0 & 82.0 \\
\textit{- Coordinates (Spatial)} & 96.3 & 57.8 & 45.7 & 40.7 & 56.7 & 54.3 & 85.2 & 72.9 \\
\textit{- Full (Times. + Coord.)} & 96.4 & 66.9 & 57.6 & 55.3 & 67.8 & 69.1 & 91.6 & 84.1 \\

\bottomrule
\end{tabular}
\end{threeparttable}
\end{table}

As shown in Table \ref{tab:internvl_ablation}, the extended experiments on InternVL3.5-8B strongly corroborate the findings from the main paper. The dense \textit{Timeseries} representation (and the \textit{Full} combination) remains the most effective grounding format, significantly outperforming the high-level \textit{Summary} embedding, particularly in physical consistency (WPCR). Furthermore, forcing the model to rely strictly on \textit{Coordinates} consistently results in a sharp performance regression across all metrics compared to the Timeseries format. 

This cross-architecture consistency confirms that the inability of current LLM backbones to reliably compute or understand discrete temporal derivatives (velocity, acceleration, jerk) from raw spatial waypoints is a generalized limitation of current foundation models. 

\subsection{Temporal Resolution Impact}

A potential confounding factor when evaluating dynamic physical reasoning from video is the temporal resolution of the visual input. In the main paper, we established a standard evaluation protocol for extracting $N=10$ evenly spaced frames from the 3.0-second clip window, yielding a frame rate of approximately 3.3 FPS. To determine whether the poor visual grounding performance was merely an artifact of temporal down-sampling, we conducted an ablation study using a higher frame rate.

We re-evaluated the Qwen3-VL-8B model on the complete benchmark using 30 frames per clip (10 FPS), effectively tripling the temporal density of the visual context.

\begin{table}[!ht]
\centering
\caption{Ablation on temporal resolution. Increasing the input frame rate from 3.3 FPS (10 frames) to 10 FPS (30 frames) for Qwen3-VL-8B yields negligible improvements in semantic ego-motion understanding. This supports the conclusion that the observed perception bottleneck stems from a fundamental representational gap, rather than insufficient temporal sampling.}
\label{tab:frame-rate-comparison}
\begin{threeparttable} 
\scriptsize
\renewcommand{\arraystretch}{1.05}
\setlength{\tabcolsep}{2pt}

\begin{tabular}{@{} l c c c c c c c c @{}} 
\toprule

\multirow{2}{*}{\textbf{Model}} 
& \textbf{Parsable} 
& \multicolumn{3}{c}{\textbf{Semantic}} 
& \multicolumn{2}{c}{\textbf{Temporal}} 
& \multicolumn{2}{c}{\textbf{Consistency}} \\

\cmidrule(lr){2-2} \cmidrule(lr){3-5} \cmidrule(lr){6-7} \cmidrule(lr){8-9}

& \textbf{$\uparrow$} 
& \textbf{Acc. $\uparrow$} 
& \textbf{BAcc. $\uparrow$} 
& \textbf{F1 $\uparrow$} 
& \textbf{Acc. $\uparrow$} 
& \textbf{F1 $\uparrow$} 
& \textbf{WPCR $\uparrow$} 
& \textbf{PCov. $\uparrow$} \\

\midrule

\rowcolor{SectionGray}
\multicolumn{9}{c}{\textbf{Qwen3-VL-8B (Main Paper, 3.3 FPS)}} \\ 
\midrule
\textit{Qwen3-VL-8B}~\cite{bai2025qwen3vltechnicalreport} & \textbf{100.0} & 49.8 & 38.9 & \textbf{33.0} & 39.3 & \textbf{32.7} & 97.9 & 84.6 \\

\midrule

\rowcolor{SectionGray}
\multicolumn{9}{c}{\textbf{Qwen3-VL-8B (10 FPS)}} \\ 
\midrule
\textit{Qwen3-VL-8B}~\cite{bai2025qwen3vltechnicalreport} & \textbf{100.0} & \textbf{50.4} & \textbf{39.3} & 32.0 & \textbf{41.7} & 32.2 & \textbf{98.5} & \textbf{84.8} \\

\bottomrule
\end{tabular}
\end{threeparttable} 
\end{table}

As detailed in~\Cref{tab:frame-rate-comparison}, tripling the frame rate yields only marginal performance deviations. The Balanced Accuracy (BAcc) increases by only 0.4 percentage points, and the semantic Macro F1 score actually shows a slight regression ($-1.0$ pp). While there is a minor gain in Temporal Accuracy (+2.4 pp), the overall physical reasoning capabilities remain severely bottlenecked. 

These results fully support the main paper's primary findings: current vision-centric foundation models struggle to directly extract or understand complex kinematic derivatives (such as acceleration and jerk) from visual observations. Because the failure mode is rooted in a visual-dynamic representation rather than simple information loss, exponentially increasing the context window by adding more frames does not resolve the reasoning gap. We acknowledge that targeted evaluation on fast-maneuver subsets at higher temporal resolutions (e.g., $\geq$30 FPS) remains an open direction, particularly as models begin to demonstrably leverage visual input for dynamic reasoning. We consider this a natural avenue for follow-up work once the underlying visual grounding deficit identified by \textit{EgoDyn-Bench} is addressed.

\section{Project Assets \& Reproducibility}
\subsection{Code and Dataset Access}
\textbf{Public release.} The complete source code, evaluation harness, baselines, and reproduction script are publicly available at \url{https://github.com/TUM-AVS/EgoDyn-Bench}. It includes:

\begin{itemize}
    \item \textbf{Dataset generation:} Labeling rules and question-answer pair generation from nuScenes and CARLA logs.
    \item \textbf{Question-answer pairs:} Ground-truth QA pairs as well as the list of selected clips for this benchmark.
    \item \textbf{Clip viewer:} The tool used to perform human-in-the-loop evaluation.
    \item \textbf{Evaluation pipeline:} Parser and metrics for benchmarking VLM responses, with batch evaluation support for multiple model providers.
    \item \textbf{Reproduction scripts:} Instructions to reproduce all reported results, and all evaluated model answers are included in the archive.
\end{itemize}
The full dataset and repository are published in accordance with the ECCV Dataset Release Policy.

\subsection{Interactive Human-in-the-Loop Evaluation Tool}
\label{sec:eval_tool}

\begin{figure}[ht]
    \centering
    \includegraphics[width=\linewidth]{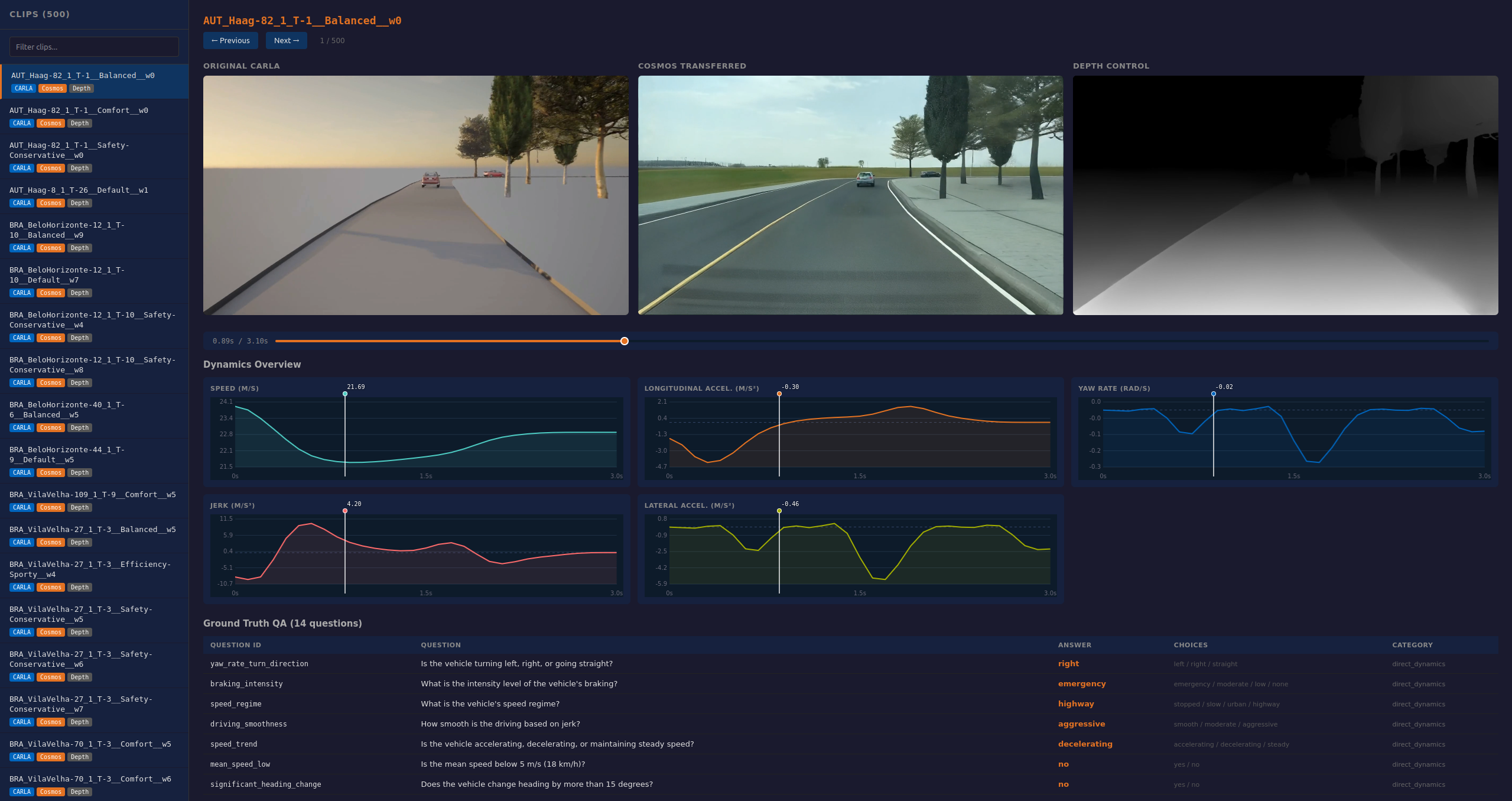}
    \caption{\textbf{Clip Viewer Web Interface.} The dashboard provides a holistic view of each benchmark sample, merging multi-modal video playback (top row), dynamic physical state tracking (middle row), and linguistic QA pairs (bottom row) into a single, synchronized timeline for human-in-the-loop verification.}
    \label{fig:clip_viewer}
\end{figure}

To ensure the high quality and precise alignment of the \textit{EgoDyn-Bench} dataset, we developed a comprehensive web-based evaluation tool. As shown in \cref{fig:clip_viewer}, this interface allows humans to verify the temporal and semantic alignment between the visual scene, physical vehicle dynamics, and the generated question-answer pairs. The clip viewer source code is part of the public release (\texttt{scripts/clip\_viewer.py}). A live interactive demo is hosted on the project page.

The tool provides the following core capabilities, designed to facilitate efficient, multi-modal data inspection:

\begin{itemize}
    \item \textbf{Synchronized Multi-Modal Playback:} The interface supports side-by-side, synchronized playback of up to three video streams per clip: the original simulator output (CARLA), the generative video transfer (Cosmos), and the depth control map. Playback features include auto-looping, a global timeline, and click-to-seek functionality.
    
    \item \textbf{Coupled Dynamics Dashboard:} A globally synchronized time bar links the video playback directly to high-resolution time-series plots. Five key dynamic state variables are visualized: speed, longitudinal acceleration, yaw rate, jerk, and lateral acceleration. The charts feature an adaptive grid layout, zero lines, and interpolated cursor readouts that update in real-time as the video plays.
    
    \item \textbf{Ground-Truth Validation:} The interface integrates the dataset's metadata, displaying computed semantic features and ground-truth QA pairs in a dedicated table. This allows reviewers to instantly cross-reference the generated text targets with the vehicle's visual and dynamic states.
    
    \item \textbf{Robust Server Backend \& Processing:} The tool is supported by a custom backend that handles on-the-fly data processing. This includes automatic transcoding of original FMP4 videos to H.264 (with caching for instant replay), an API for extracting evenly-spaced frame montages, and a timeseries API that serves array data and computes lateral acceleration dynamically (as $\text{speed} \times \text{yaw\_rate}$).
\end{itemize}

\subsection{Project Page and Public Release}
\label{sec:project_page}
All artifacts are tagged at release \texttt{v1.0}. The evaluation harness, baselines, and reproduction scripts are available at \url{https://github.com/TUM-AVS/EgoDyn-Bench} under the Apache 2.0 license. The dataset is published at \url{https://huggingface.co/datasets/fnc1901/EgoDyn-Bench} under CC BY-NC-SA 4.0, chosen to comply with nuScenes' upstream license. It contains:
\begin{itemize}
    \item the canonical 1{,}000-clip benchmark spec and oracle QA,
    \item per-clip dynamics arrays at 31 samples $\times$ 7 channels,
    \item both CARLA visual domains (raw simulation and Cosmos-Transferred),
    \item the 49-model reference leaderboard together with raw model outputs, and
    \item the 80-clip visual-artifact subset list discussed in Sec.~\ref{subsec:domain_gap}.
\end{itemize}
Raw nuScenes imagery is not redistributed. Users obtain nuScenes separately from \url{https://www.nuscenes.org} and join with the released dataset via the included \texttt{sample\_token} references.
The additional project page features the following core components:

\begin{itemize}
    \item \textbf{Comprehensive Leaderboard:} A public, dynamic ranking of all evaluated VLMs on the benchmark, reporting all metrics from the main paper.
    
    \item \textbf{Granular Performance Analysis:} 
    \begin{itemize}
        \item \textit{Per-Question Type:} Detailed performance breakdowns across all 14 question categories, allowing researchers to pinpoint specific kinematic reasoning deficits (e.g., speed vs. yaw rate) for each VLM.
        \item \textit{Source-Level Domain Gap:} Separate result stratifications for real-world (\textit{nuScenes}) versus simulated (\textit{CARLA}) clips to analyze the sim-to-real domain gap in VLM video understanding.
    \end{itemize}
    
    \item \textbf{Dynamics Embedding Ablations:} Interactive visualizations highlighting the performance delta between video-only baselines and models augmented with textual dynamics embeddings. This allows users to easily identify which specific question types benefit most from explicit dynamic state information.
    
    \item \textbf{Dataset Browser and Interactive Demo:} As detailed in \cref{sec:eval_tool}, the project page includes a fully functional dataset explorer. Users can search and filter the clip list, view computed kinematic-feature badges, and use our synchronized multimodal clip viewer to inspect ground-truth QA pairs alongside the video and time-series data.
\end{itemize}

%
%
\section*{Acknowledgment}
This research was conducted in collaboration with BMW Group and was supported by their research funding. Generative AI tools were used for language editing and proofreading during manuscript preparation. All content was reviewed and verified by the authors, who take full responsibility for the final manuscript.

\end{document}